\documentclass[10pt,twocolumn,letterpaper]{article}

\usepackage{cvpr}              

\definecolor{darkgreen}{rgb}{0.0, 0.5, 0.0}
\definecolor{darkred}{rgb}{0.5, 0.0, 0.0}

\newcommand{\red}[1]{{\color{darkred}#1}}
\newcommand{\green}[1]{{\color{darkgreen}#1}}

\usepackage{multirow}
\usepackage{threeparttable}
\usepackage{algorithm}
\usepackage{algpseudocode}
\usepackage{makecell}

\definecolor{cvprblue}{rgb}{0.21,0.49,0.74}
\usepackage[pagebackref,breaklinks,colorlinks,allcolors=cvprblue]{hyperref}

\def\paperID{*****} 
\def\confName{CVPR}
\def\confYear{2026}

\title{Enhancing Facial Expression Recognition in Head-Mounted Displays with Synthetic Data}

\author{Jianing Deng \quad Qiang Zhou \quad Jingtong Hu\\
University of Pittsburgh\\
}

\begin{document}
\maketitle
\begin{abstract}
Facial expression recognition (FER) is crucial for social interaction in mixed reality environments that employ head-mounted displays (HMD). However, collecting FER data from head-mounted cameras (HMC) is challenging due to privacy concerns and the diversity of HMD platforms. Moreover, existing FER datasets are not directly applicable due to the unique perspectives of HMCs. The lack of sufficient data hinders the development of neural network-based HMC FER methods. To address data scarcity, we propose a data synthesis framework that generates HMC-view images from frontal-view images, leveraging abundant existing annotated datasets. Specifically, we first reconstruct 3D textured meshes from images and then apply a configurable camera system to render images from the HMC perspective. Additionally, we introduce a texture-space alignment network (TSAN) that enables accurate texture sampling from images to preserve detailed facial expressions. To evaluate the proposed method, we conduct extensive experiments on both simulated and real HMC datasets. Experimental results demonstrate that models trained on our synthetic dataset outperform those trained on existing datasets and exhibit better generalization across different camera configurations.
\end{abstract}

\section{Introduction}
\label{sec:intro}

Mixed reality (MR) has experienced rapid advancements in recent years, with the potential to become an integral part of daily life~\cite{speicher2019mixed,grandview2024mr}. Recognizing facial expressions is crucial for daily interactions, facilitating both personal emotion tracking and effective interpersonal communication. However, the use of head-mounted displays (HMD) in MR devices presents challenges for facial expression recognition (FER) due to the obstruction of large portions of the face. To address this issue, previous work~\cite{lombardi2018deep,thies2018facevr,wei2019vr,hickson2019eyemotion,jourabloo2022robust,martinezcodec} leverages \emph{user-facing} head-mounted cameras (HMC) embedded within HMD to capture detailed motions around the eyes and mouth. An example of HMC-view images is illustrated in~\cref{tab1} (b).

\begin{table}
  \centering
  \small
  \begin{tabular}{ll}
    \toprule
    (a) Traditional FER Dataset & (b) HMC FER Dataset \label{item:B} \\
    
    \begin{minipage}{0.45\linewidth}
        \centering
        \includegraphics[width=0.48\linewidth]{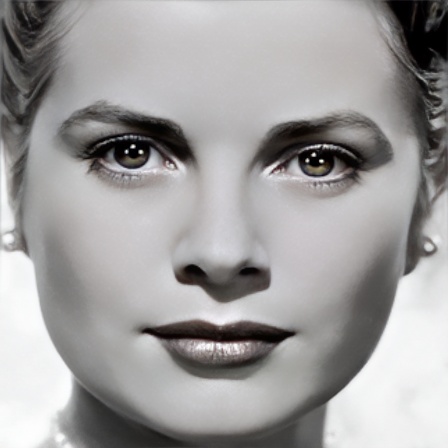}
        \includegraphics[width=0.48\linewidth]{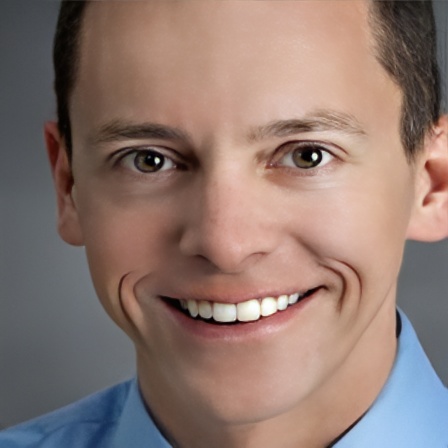}
    \end{minipage}  &
    \begin{minipage}{0.45\linewidth}
        \centering
        \includegraphics[width=0.48\linewidth]{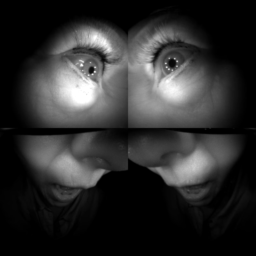}
        \includegraphics[width=0.48\linewidth]{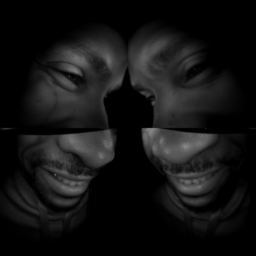}
    \end{minipage} \\
    
    \addlinespace[5pt]
    
    \begin{minipage}{0.45\linewidth}
        \begin{itemize}
            \item Frontal-view head portrait
            \item RGB
            \item \green{Millions of images}
            \item \red{Unsuitable for HMC FER}
        \end{itemize}
    \end{minipage} &
    \begin{minipage}{0.45\linewidth}
        \begin{itemize}
            \item HMC close-up view
            \item Infrared (IR)
            \item \red{256 subjects (4K samples)}
            \item \red{Limited HMC settings}
        \end{itemize}
    \end{minipage} \\

    \addlinespace[4pt]
    \toprule
    \multicolumn{2}{l}{(c) Our Synthetic HMC Dataset} \\
    \begin{minipage}{0.45\linewidth}
        \centering
        \includegraphics[width=0.48\linewidth]{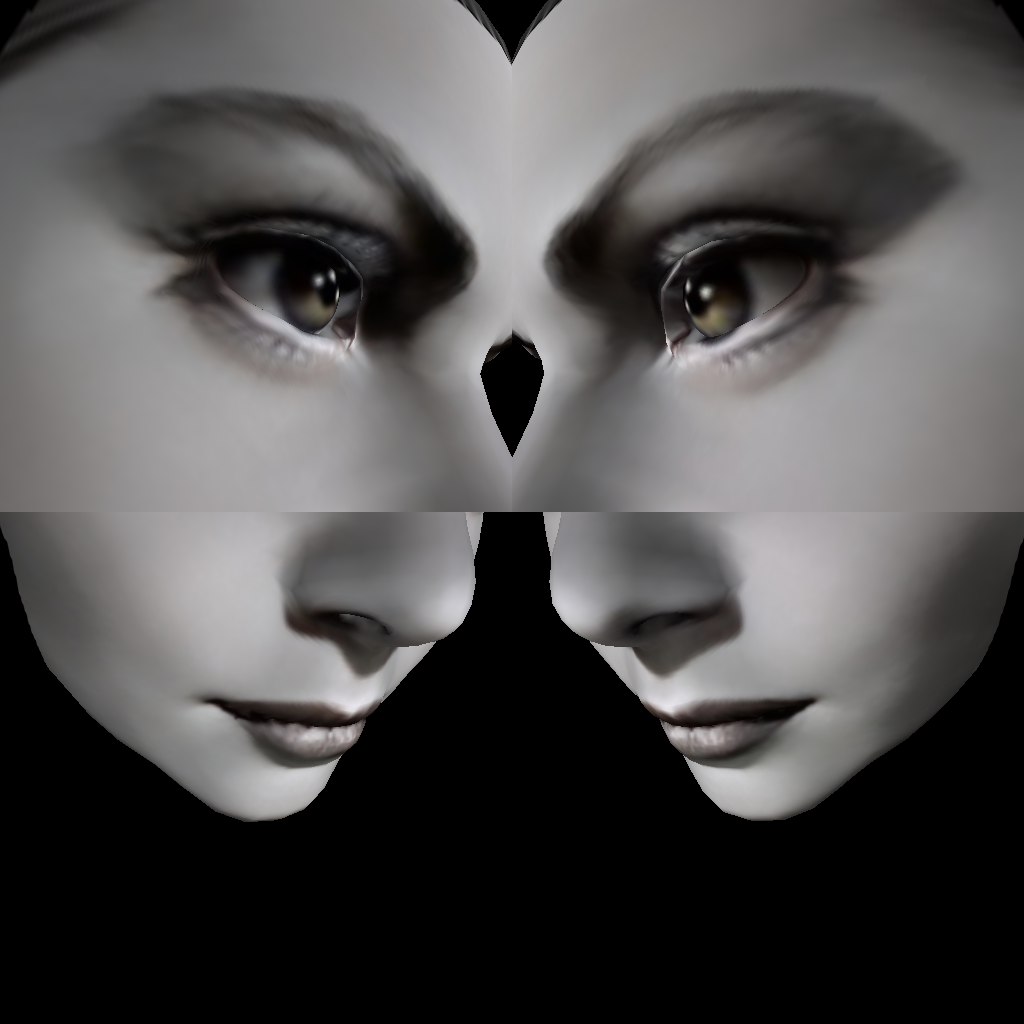}
        \includegraphics[width=0.48\linewidth]{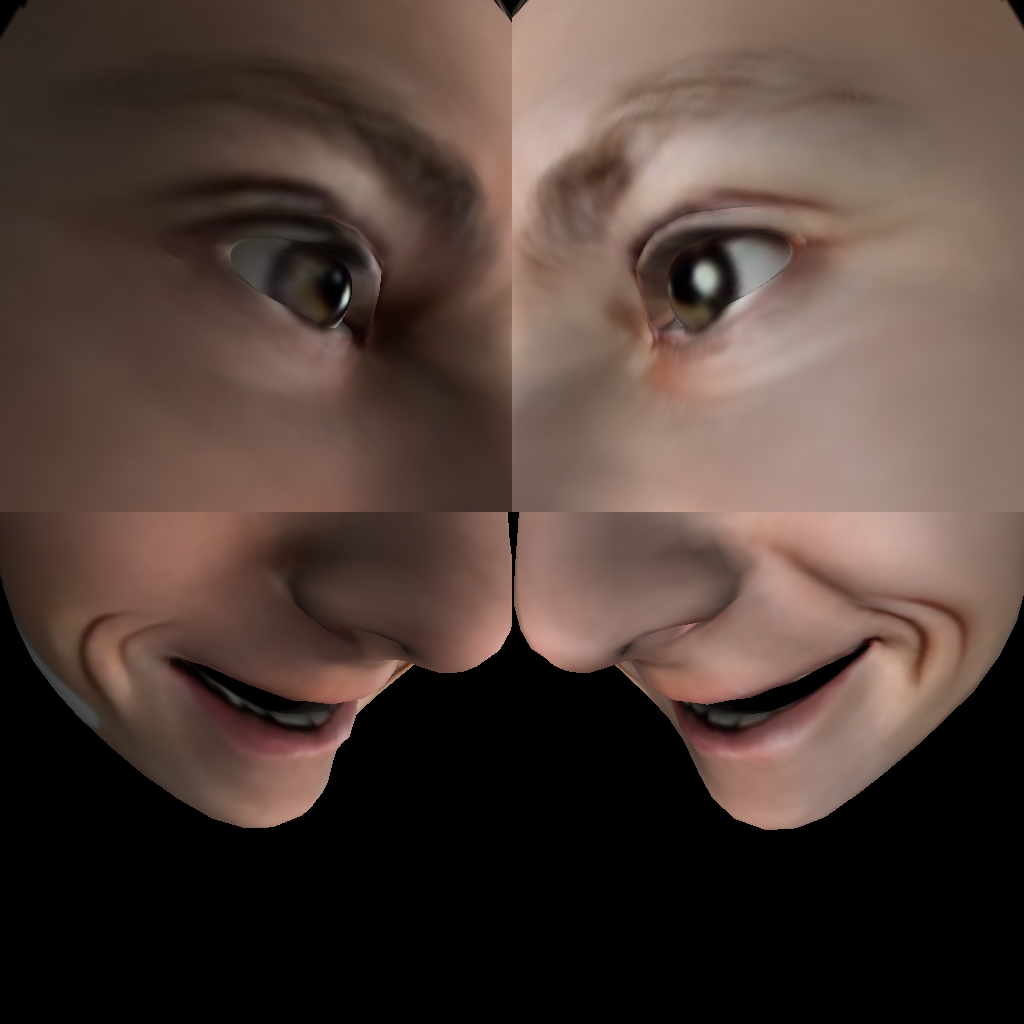}
    \end{minipage} & 
    \begin{minipage}{0.45\linewidth}
        \begin{itemize}
            \item HMC close-up view
            \item RGB
            \item \green{Millions of samples}
            \item \green{Diverse HMC settings}
        \end{itemize}
    \end{minipage} \\

    \bottomrule
  \end{tabular}
  \caption{
    \label{tab1}
    Comparison of publicly available datasets for (a) traditional FER~\cite{mollahosseini2017affectnet} and (b) HMC FER~\cite{martinezcodec}. (c) Our method addresses data scarcity in HMC FER by synthesizing millions of photo-realistic HMC-view samples for model training.
  }
\end{table}

One of the main challenges in HMC FER is data scarcity, as real-world datasets are limited due to labor-intensive and costly collection~\cite{martinezcodec}, privacy concerns, and device variability across HMD platforms.  Hickson \etal~\cite{hickson2019eyemotion} curated an in-house FER dataset with 46 subjects using eye-tracking cameras. While effective in estimating expressions around the eyes, it fails to capture expressions in other facial areas, such as mouth movements.  
Martinez \etal~\cite{martinezcodec} collected an HMC dataset using a commercial headset. While it is the largest publicly available dataset for HMC FER, it is limited to 256 subjects with a fixed HMC configuration.
The limited availability and lack of diversity in HMC data significantly hinder the development of machine learning methods, especially deep neural networks.

In contrast, traditional FER using (near) frontal-view (FV) face images has been well-studied, with millions of labeled samples from controlled lab settings and online sources~\cite{mollahosseini2017affectnet,li2017reliable}.
A straightforward approach is to train models on FV FER datasets and apply them to HMC data. However, we observed that these models perform poorly on HMC data, indicating a significant \emph{domain gap} between FV and HMC FER. 
We attribute this primarily to substantial differences in \emph{camera perspective}.
A detailed analysis is provided in \cref{sec:motivation}.

To harness the abundant FER datasets while addressing the aforementioned issues, we propose a data synthesis framework that transforms frontal-view images into HMC-view counterparts. Our framework first reconstructs an expressive 3D mesh from each image and then adopts a virtual headset system to render HMC images from the mesh.
Specifically, we leverage existing face reconstruction techniques~\cite{feng2021learning,danvevcek2022emoca} for mesh geometry estimation.
For mesh texture estimation, we found that existing PCA-based linear methods~\cite{paysan20093d,feng2021learning} and generative methods~\cite{lattas2020avatarme,bai2023ffhq} fail to effectively capture fine-grained expression details, such as wrinkles and furrows, which are critical for expression recognition~\cite{tian2001recognizing}.
To address this, we adopt a texture sampling approach from the frontal image to better preserve fine details. To ensure accurate sampling, we further introduce a novel Texture-Space Alignment Network (TSAN), which learns a rectified flow field to correct misaligned textures in texture space.

To evaluate the proposed method, we conduct extensive experiments on both simulated and real HMC datasets. The results show that models trained on our synthetic dataset outperform those trained on existing datasets and generalize better across different camera configurations.

\paragraph{Contributions.} This work pioneers the use of synthetic data for HMC FER and demonstrates its effectiveness in addressing data scarcity.
We identify the challenge of limited cross-view generalization in neural networks and introduce a novel data synthesis pipeline featuring two key innovations: (1) a direct texture sampling strategy to generate diverse and detailed facial expressions, and (2) a texture alignment method that mitigates artifacts in reconstructed textures, enhancing FER performance.
Furthermore, we introduce a comprehensive benchmark for HMC FER by incorporating both simulated and real HMC datasets. Evaluation results show that models trained on our synthetic dataset consistently outperform those trained on existing datasets.
Our work enables the reuse of large-scale annotated frontal FER data and paves the way for future research in HMC FER.


\section{Related Work}

\subsection{Facial Expression Recognition}

Facial expression recognition (FER) has been an active research area for decades~\cite{ekman1992argument}. Traditional FER research primarily focuses on recognizing expressions from facial portrait images.
In recent years, deep neural network (DNN)-based FER methods~\cite{wang2020region,wang2020suppressing,zhao2021robust,xue2022vision} have achieved state-of-the-art performance, with research gradually shifting toward addressing real-world challenges such as occlusion~\cite{lee2023latent}, pose variation~\cite{jampour2022multiview}, label ambiguity~\cite{wang2020suppressing,zhang2022learn}, and domain gap~\cite{zhang2024generalizable,zhao2025enhancing}.
Another line of work explores dynamic FER~\cite{zhao2021former,wang2023rethinking}, which leverages videos instead of static images to incorporate temporal cues for improved FER.

Despite significant progress in FER, limited research has focused on recognizing facial expressions from egocentric HMC views, primarily due to the scarcity of training and evaluation datasets.
The most relevant work is by Hickson~\etal~\cite{hickson2019eyemotion}, who curated an in-house dataset using two eye-tracking cameras and successfully trained a personalized FER model on the dataset. However, the dataset is not publicly available and is restricted to capturing only the eye region, neglecting other facial areas crucial for comprehensive expression recognition. Martinez \etal~\cite{martinezcodec} recently released Ava256, a dataset captured with a commercial headset covering upper and lower facial regions across diverse expressions. However, the limited amount of data makes it unsuitable for \emph{training} DNN models.

\subsection{Monocular 3D Face Reconstruction}

Reconstructing a 3D face from a single image has received significant attention in recent years. Given the highly ill-posed nature of the task, we focus on 3D Morphable Model (3DMM)–based methods~\cite{blanz1999morphable,li2017learning,wood2021fake,bao2021high} for their robustness. The reconstruction process typically consists of two components: shape reconstruction, which determines the 3D vertex positions of the face mesh, and texture reconstruction, which computes per-vertex colors or a texture atlas for shading.
For shape reconstruction, existing methods either directly regress 3DMM geometry parameters~\cite{deng2019accurate,feng2021learning,danvevcek2022emoca} or predict dense facial landmarks~\cite{feng2018joint,wood20223d,taubner20243d} from images.
For texture reconstruction, prior works~\cite{blanz1999morphable,deng2019accurate,feng2021learning,danvevcek2022emoca} relied on PCA-based albedo models that often produced blurry and unrealistic textures, while more recent approaches use GANs to generate more realistic facial textures~\cite{lattas2023fitme,rai2024towards,bai2023ffhq}.
Only a few studies~\cite{chen2019photo,rai2024towards} have attempted to unwrap textures directly from images. However, their performance is limited due to the challenge of achieving precise alignment between mesh and image.

\subsection{Synthetic Data for Training Neural Networks}

Data scarcity and imbalance present significant challenges in training deep learning models. Recently, synthetic data has been widely adopted as a solution to data scarcity across low- and high-level vision tasks~\cite{raistrick2023infinite,wu2023synthetic}, robotics~\cite{Genesis}, healthcare~\cite{mcduff2023synthetic}, and other domains. For a comprehensive review of synthetic data in computer vision and related tasks, see~\cite{nikolenko2021synthetic,paproki2024synthetic,joshi2024synthetic}.
More relevant to this work,  Wei~\etal~\cite{wei2019vr,jourabloo2022robust} synthesized HMC-mesh paired data to enable 3D head avatar reconstruction from HMC images.
Wood~\etal~\cite{wood2021fake,wood20223d} demonstrated the feasibility of using synthetic data for mid-level face-related vision tasks such as face segmentation and landmark detection. However, their synthetic data has not been shown effective for FER which require high-level semantic understanding. Moreover, they use a limited set of manually cleaned texture maps for face generation, which lack diversity and realistic expression details. In contrast, we sample diverse textures with rich expression details from images, making our approach better suited for HMC FER.

\section{Domain Gap: From Frontal to HMC FER}
\label{sec:motivation}

Previous study~\cite{zhang2024generalizable} has identified a domain gap among traditional frontal-view (FV) FER datasets. In this section, we examine whether this gap is further exacerbated when shifting from FV datasets to HMC datasets.

To this end, we train models using AffectNet~\cite{mollahosseini2017affectnet}, an FV dataset, and evaluate them on both the FV data and the HMC data simulated from 3DRFE~\cite{stratou2011effect, ma2007rapid}.
Specifically, we train two models, one using ResNet~\cite{he2016deep} and the other using Swin Transformer~\cite{liu2021swin}, on the AffectNet dataset for seven-class basic expression classification.

\begin{figure}[t]
    \centering
    \includegraphics[width=\columnwidth]{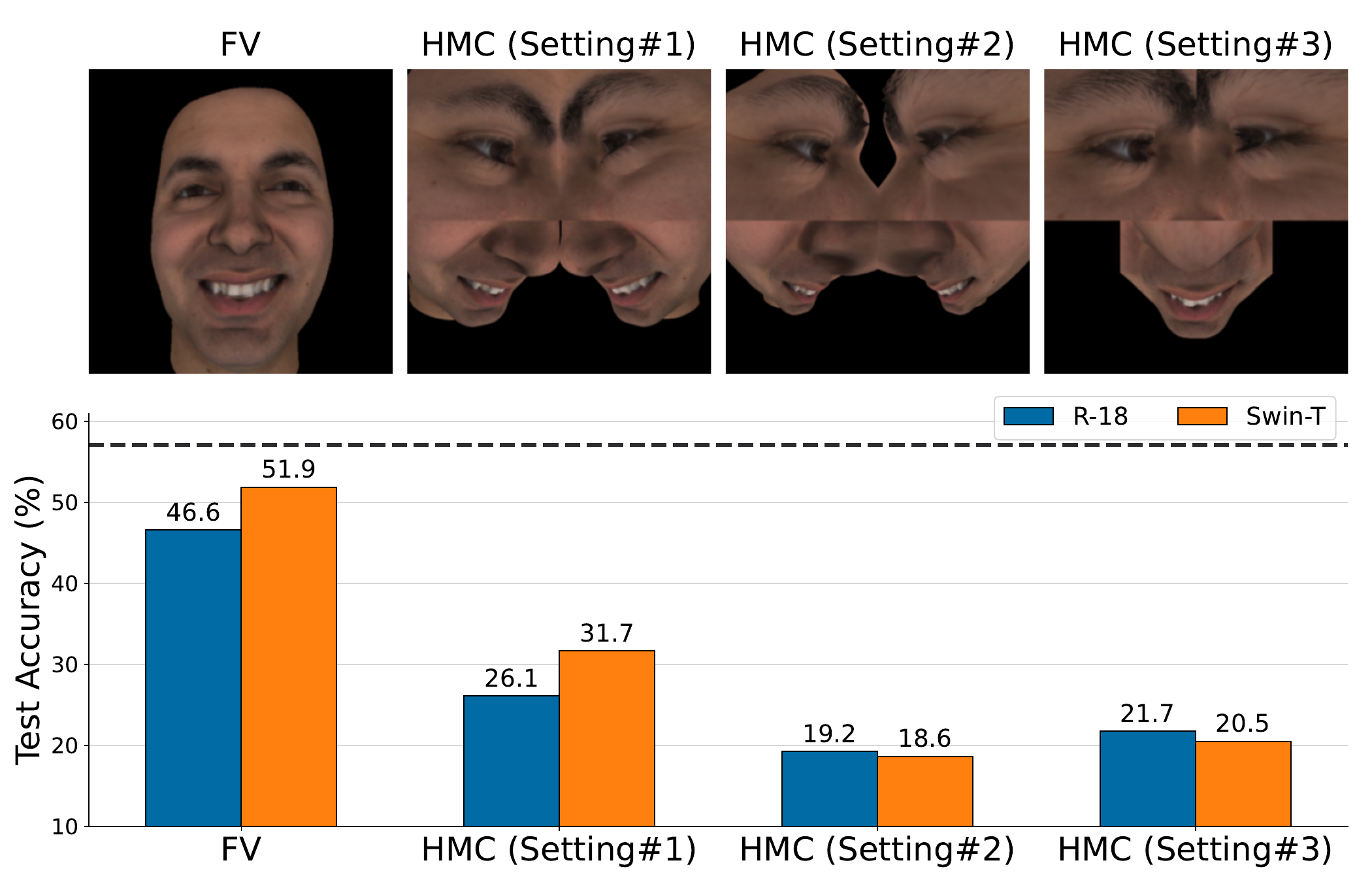}

    \vspace{-5pt}
    \caption{Accuracy of \emph{AffectNet-trained} models evaluated on the 3DRFE dataset with different \emph{input sources}. The first row visualizes the input data fed into the networks. The dashed line represents the accuracy of the Swin-T model on the AffectNet test set.
    }
    \label{fig:motivation}
\end{figure}

As illustrated in \cref{fig:motivation}, we conducted four sets of experiments to evaluate the two trained models.
In the first experiment, the models are tested on the FV dataset rendered from 3DRFE, while in the remaining three experiments, they are evaluated on the HMC dataset rendered from 3DRFE, each with a specific camera setting.
Specifically, Setting\#1 represents a relatively normal viewpoint with four cameras, Setting\#2 features a wide field of view (FoV) with the same number of cameras, and Setting\#3 maintains a normal viewpoint but with three cameras. Based on the experimental results, we have the following observations:

\begin{itemize}

    \item \textbf{A minor domain gap exists between the two FV datasets.} Compared to the model's performance on the source domain (\ie, the AffectNet test set, represented by the dashed line in~\cref{fig:motivation}), only a slight drop in accuracy is observed when tested on the 3DRFE FV dataset.

    \item \textbf{A large domain gap exists between the FV and HMC datasets.} When tested on HMC Setting\#1, accuracy drops significantly by 44\% for ResNet-18 and 39\% for Swin-T, indicating that models trained on FV data struggle with HMC data due to significant viewpoint discrepancies.

    \item \textbf{Model performance varies across different HMC settings.}  In Setting\#2, the wider FoV and increased perspective distortion result in an average 61\% drop in test accuracy. In Setting\#3, the use of three camera angles leads to a averaged 57\% decline in test accuracy.
    
\end{itemize}

\begin{figure*}[t]
    \centering
    \includegraphics[draft=false, width=\linewidth]{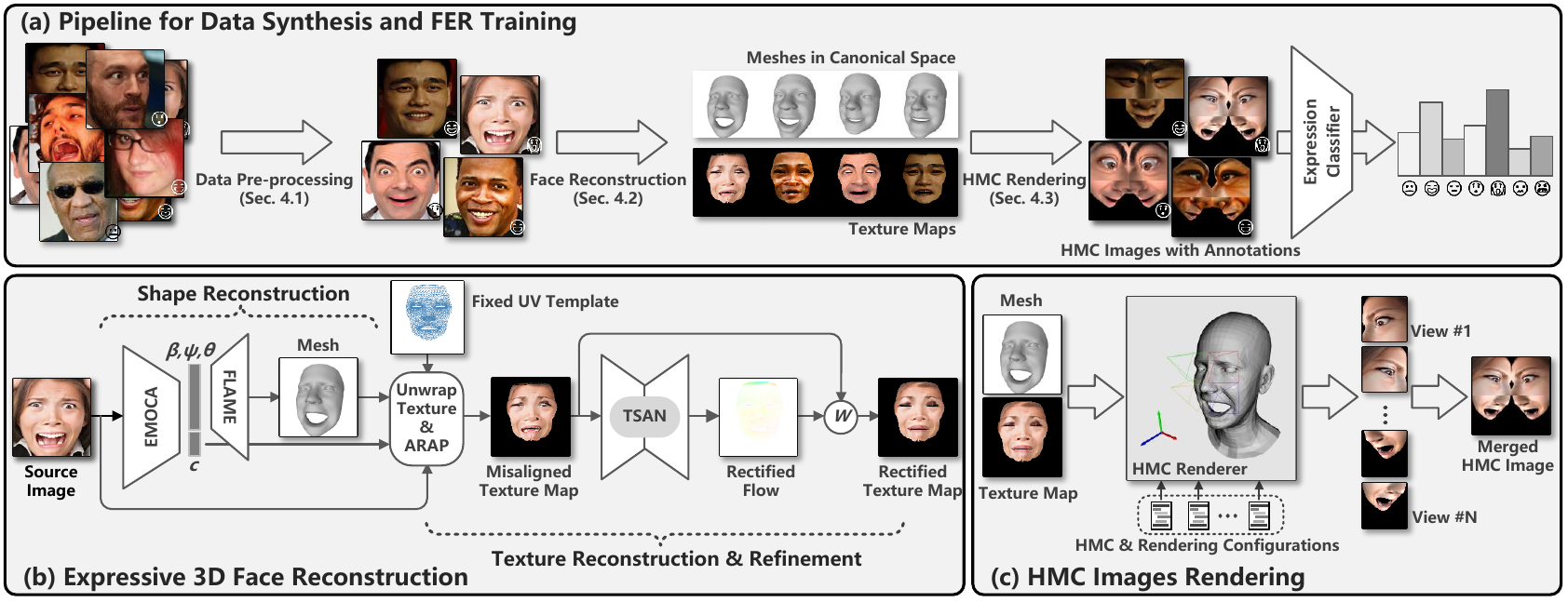}

    \vspace{-5pt}
    \caption{Overview of the proposed \textbf{SynHMC} framework for synthetic HMC images generation and HMC FER training.}
    \label{fig:overview}
\end{figure*}

These findings indicate that training solely on existing FV datasets is insufficient for models to generalize to HMC images, especially in practical scenarios with diverse camera configurations. This highlights the need for a dedicated HMC dataset that accounts for such variations.
\section{Data Synthesis Framework for HMC FER}

\subsection{Overview}

The objective of HMC FER is to classify HMC images into one of seven basic expressions~\cite{ekman1992argument,mollahosseini2017affectnet}: neutral, happy, sad, surprised, fearful, disgusted, or angry. Given the scarcity of HMC-view data and the domain gap shown in the previous section, we propose synthesizing HMC-view counterparts from frontal images to facilitate model training.

\Cref{fig:overview} presents an overview of the proposed SynHMC framework. We begin with data pre-processing to exclude samples that are unsuitable for face reconstruction. Specifically, we filter out: 1) images where subjects wear glasses, detected using a face parsing model~\cite{zllrunningfaceparsing}; and 2) images with excessive yaw angles, determined using the estimated 3DMM pose.
Next, we reconstruct a textured mesh for each image using the proposed 3D reconstruction method in \cref{sec:emo_3d_recon}. We then render HMC images from the reconstructed 3D faces using a configurable camera system, which can vary across different HMC settings, as described in \cref{sec:hmc_render}. Finally, we merge HMC images from all camera views into a single image and pair them with expression labels for expression classifier training.

\subsection{Expressive 3D Face Reconstruction}
\label{sec:emo_3d_recon}

To render HMC-view images, we reconstruct a 3D face from a single frontal-view image while preserving expression details such as wrinkles and furrows. For geometry modeling, given the highly ill-posed nature of the problem, we employ the statistical 3D Morphable Model (3DMM)~\cite{blanz1999morphable}, which provides strong prior information for robust face reconstruction.
For appearance modeling, previous PCA-based linear methods~\cite{paysan20093d,feng2021learning} and generative methods~\cite{lattas2020avatarme,bai2023ffhq} struggle to capture expression details and diverse appearance changes. To address this, we \emph{directly sample pixels} (\ie, unwrapping) from source images, preserving expression details as much as possible.

\noindent\textbf{Shape reconstruction.}
We employ FLAME~\cite{li2017learning} to generate face mesh, which is a 3DMM that parameterizes facial shape, expression, and pose. The identity shape is controlled by parameters $\boldsymbol{\beta} \in \mathbb{R}^{|\boldsymbol{\beta}|}$, while facial expressions are represented by $\boldsymbol{\psi} \in \mathbb{R}^{|\boldsymbol{\psi}|}$. The pose parameters $\boldsymbol{\theta} \in \mathbb{R}^{3k+3}$ define the rotations of $k = 4$ joints (neck, jaw, and eyeballs) along with the global rotation. Given these parameters, FLAME generates a 3D face mesh as
\vspace{-5pt}
\begin{equation}
M(\boldsymbol{\beta}, \boldsymbol{\psi}, \boldsymbol{\theta}) \to (\mathbf{V}, \mathbf{F})
\vspace{-5pt}
\end{equation}
where the vertex positions are given by $\mathbf{V} \in \mathbb{R}^{N_v \times 3}$ with $N_v = 5023$ vertices and the connectivity is described by $\mathbf{F} \in \mathbb{R}^{N_f \times 3}$ with $N_f = 9976$ faces.

To obtain the aforementioned FLAME parameters, we utilize EMOCA~\cite{danvevcek2022emoca}, a regression-based face tracking method, to estimate them from the source image $\mathbf{I}_\mathrm{src} \in \mathbb{R}^{H \times W \times 3}$. EMOCA also estimates the orthographic camera parameters $\boldsymbol{c} \in \mathbb{R}^3$, allowing reprojection of the 3D face mesh into image space.

\noindent\textbf{Texture reconstruction.}
Given the estimated mesh and camera, we compute the mapping from any point $\mathbf{t} = (u, v)$ in the texture (UV) space $\mathbf{I}_\mathrm{tex} \in \mathbb{R}^{H \times W \times 3}$ to its corresponding point $\mathbf{p} = (x,y)$ in the source image space, followed by sampling the texture color from the image. This process is formulated as:
\begin{equation}
\mathbf{I}_\mathrm{tex}(\mathbf{t}) = \mathcal{I}(\mathbf{I}_\mathrm{src}, \mathbf{p}),  \quad 
\mathbf{p} = \mathcal{P} \circ \mathcal{M} (\mathbf{t}).
\vspace{-5pt}
\end{equation}
where $\mathcal{I}$ denotes bilinear interpolation, $\mathcal{P}$ is the projection function that maps the 3D point onto the image plane using camera $\boldsymbol{c}$, and $\mathcal{M}$ is the mapping function from texture space to 3D space via barycentric interpolation based on triangle correspondences. 
For small holes in the unwrapped $\mathbf{I}_\mathrm{tex}$ caused by self-occlusion, we fill them using a Fast Marching-based method~\cite{telea2004image} for seamless texture.

Existing FER datasets typically contain low-resolution images with compression artifacts. To enhance source image quality, we apply GFP-GAN~\cite{wang2021towards} to remove artifacts and upscale resolution by $2\times$ before unwrapping. We found that GFP-GAN effectively preserves expression details.

\begin{figure}[t]
    \centering
    \small
    \subcaptionbox*{\scriptsize Source Image}{\includegraphics[width=0.243\linewidth]{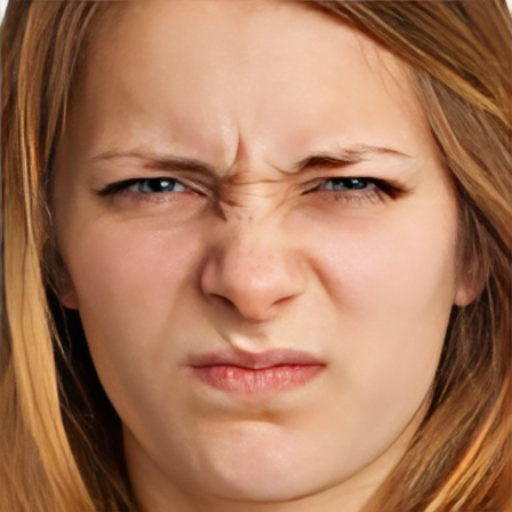}}
    \subcaptionbox*{\scriptsize Misalignment}{\includegraphics[width=0.243\linewidth]{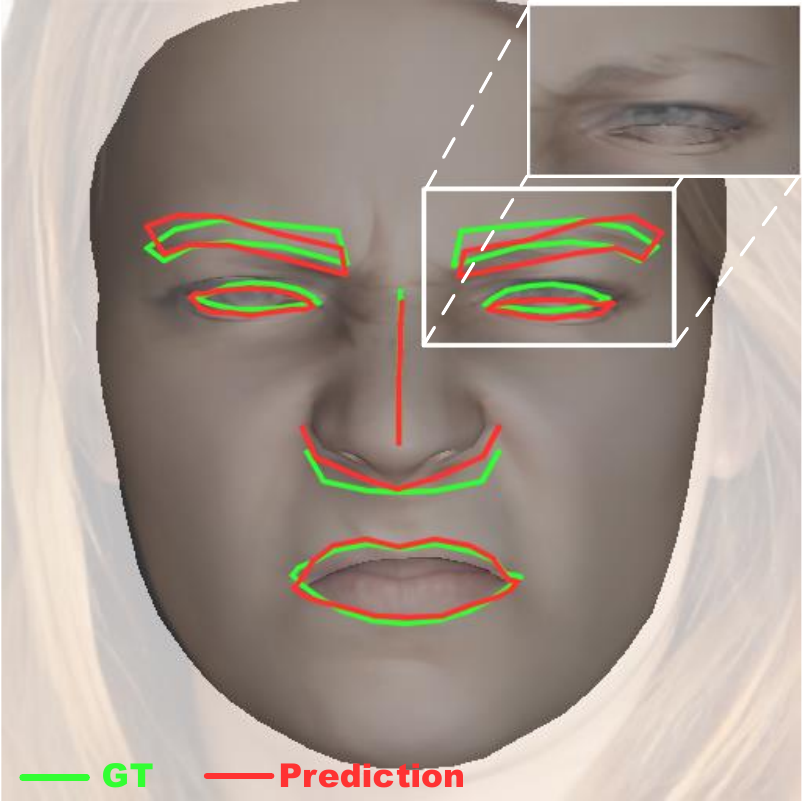}}
    \subcaptionbox*{\scriptsize Texture Map}{\includegraphics[width=0.243\linewidth]{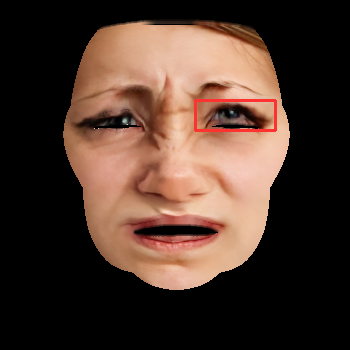}}
    \subcaptionbox*{\scriptsize HMC Image}{\includegraphics[width=0.243\linewidth]{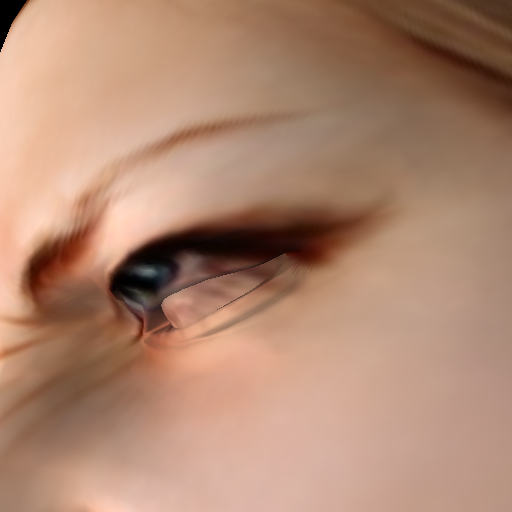}}

    \vspace{-8pt}
    \caption{Texture unwrapped using the estimated shape and camera parameters from~\cite{chu2024gpavatar}.
    The resulting texture map is misaligned and distorted, causing artifacts in rendered images.}
    \label{fig:demo_misalignment}
\end{figure}

\noindent\textbf{Texture refinement.}
Precise alignment between the projected mesh and the image is crucial for accurate texture unwrapping. However, as shown in~\cref{fig:demo_misalignment}, existing methods~\cite{deng2019accurate,feng2021learning,danvevcek2022emoca,chu2024gpavatar} often produce misalignments due to ambiguities in shape and camera pose. To address this, we follow~\cite{zhou2019deep} and apply the As-Rigid-As-Possible (ARAP)~\cite{sorkine2007rigid} transformation to adjust each point $\mathbf{p}$ using landmarks predicted by an off-the-shelf detector~\cite{lugaresi2019mediapipe}.
Nevertheless, we found that ARAP becomes ineffective when the detected landmarks are inaccurate.

\begin{figure}[t]
    \centering
    \subcaptionbox*{\scriptsize Anchor Mesh}{\includegraphics[width=0.190\linewidth]{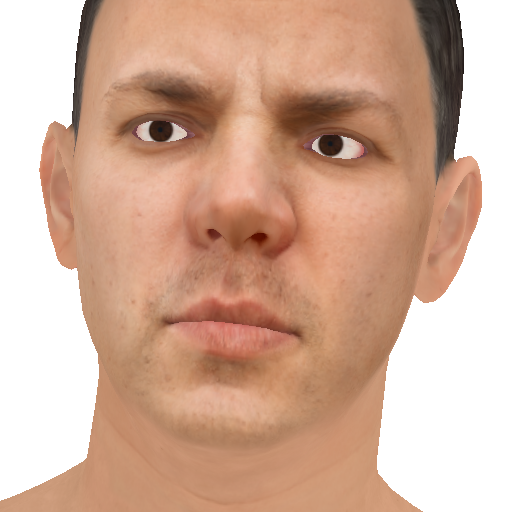}}
    \subcaptionbox*{\scriptsize Perturbed Mesh}{\includegraphics[width=0.190\linewidth]{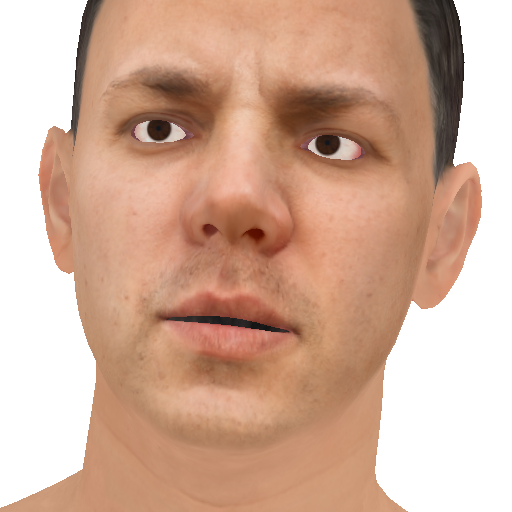}}
    \subcaptionbox*{\scriptsize Difference}{\includegraphics[width=0.190\linewidth]{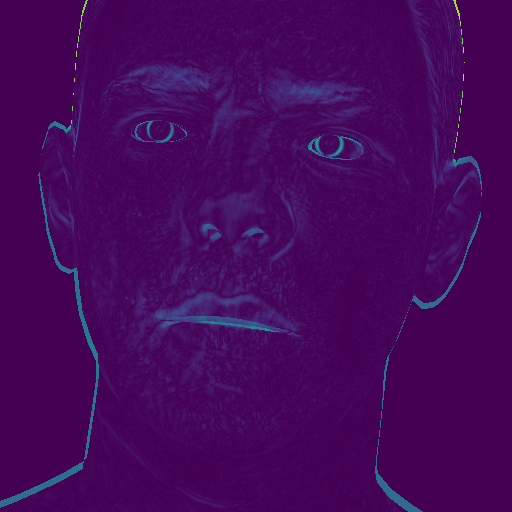}}
    \subcaptionbox*{\scriptsize Target Tex}{\includegraphics[width=0.190\linewidth]{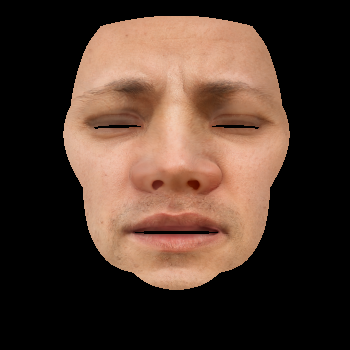}}
    \subcaptionbox*{\scriptsize Misaligned Tex}{\includegraphics[width=0.190\linewidth]{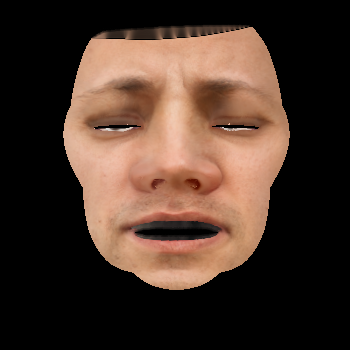}}

    \vspace{-8pt}
    \caption{Illustration of synthetic texture pair for TSAN training.}
    \label{fig:demo_uvflow_data}
\end{figure}

Rather than aligning in image space, we further tackle the misalignment problem in texture space. Our intuition is to utilize the \emph{fixed topology structure} of the UV template, \ie, each pixel in the texture space corresponds to a \emph{specific} semantic region of the face, regardless of changes in 3D vertex positions. Specifically, we propose estimating a 2D rectified flow field $\mathbf{F} \in \mathbb{R}^{H \times W \times 2}$ that maps the misaligned texture map to the UV template using a Texture-Space Alignment Network (TSAN). With the rectified flow, we can correct the misaligned texture map as:
\begin{equation}
    \mathbf{I}'_\mathrm{tex} = \mathcal{W}(\mathbf{I}_\mathrm{tex}, \mathrm{TSAN} \left( \mathbf{I}_\mathrm{tex}; \boldsymbol{\varphi}) \right)
\label{eq:4}
\vspace{-5pt}
\end{equation}
where $\mathcal{W}$ represents the backward warping~\cite{jiang2018super} operation, and $\boldsymbol{\varphi}$ denotes the learnable parameters of TSAN. We employ a simple U-Net~\cite{ronneberger2015u} architecture for TSAN, incorporating a learnable positional encoding~\cite{vaswani2017attention} map to encode positional information.

Since ground-truth rectified flow is unavailable, we train TSAN by synthesizing misaligned and ground-truth texture pairs using the estimated shape and an off-the-shelf texture map database~\cite{bai2023ffhq}. Specifically, we introduce random perturbations to shape \(\boldsymbol{\beta}\), expression \(\boldsymbol{\psi}\), and camera \(\boldsymbol{c}\) parameters to generate misaligned meshes with varied textures, then render and unwrap them to obtain misaligned texture maps. \Cref{fig:demo_uvflow_data} shows an example of the synthetic training image pair. We optimize TSAN by minimizing the L1 distance between misaligned and ground-truth image pairs. More details are provided in Appendix A1.

\subsection{HMC Images Rendering}
\label{sec:hmc_render}

Given a reconstructed 3D face mesh, this section introduces the modeling of HMD cameras and the rendering of HMC images. Our HMC rendering process is highly flexible and configurable, accommodating practical scenarios where various camera systems are used across different headsets.

\noindent\textbf{Camera system.}
We model the HMD camera system as an articulated system, similar to the head modeling in ~\cite{li2017learning}. The camera system consists of $N+1$ joints: a root joint for the headset's position and orientation, and $N$ child joints that determine the positions and orientations of the cameras. We parameterize the root joint using a \emph{global} orientation matrix $\mathbf{Q}_0 \in SO(3)$ and position vector $\mathbf{l}_0 \in \mathbb{R}^3$, while the $i$-th camera joint is defined by a \emph{relative} rotation matrix $\mathbf{R}_i \in SO(3)$ and translation vector $\mathbf{t}_i \in \mathbb{R}^3$. Symmetry is imposed between the corresponding left and right cameras. The orientation and position of the $i$-th camera in world space are determined as follows:
\begin{equation}
\mathbf{Q}_i = \mathbf{Q}_0 \mathbf{R}_i, \quad 
\mathbf{l}_i = \mathbf{l}_0 + \mathbf{Q}_0 \mathbf{t}_i.
\label{eq:4}
\vspace{-5pt}
\end{equation}

We adopt a perspective camera model parameterized by the field of view (FoV), providing a unified representation of focal length and sensor size in real-world cameras.

\noindent\textbf{Rendering.}
We employ directional lighting from the viewpoint of HMCs toward the face. Since we use textures unwrapped from lit images, which typically include facial highlights, we apply only a Lambertian diffuse term for shading and omit the specular term. We render the $N$-view HMC images simultaneously and spatially merge them into a single HMC image for FER training.

\section{Experiments}
\label{sec:experiments}

\begin{figure}[t]
    \centering
    \subcaptionbox*{\scriptsize Real HMC}{\includegraphics[width=0.158\linewidth]{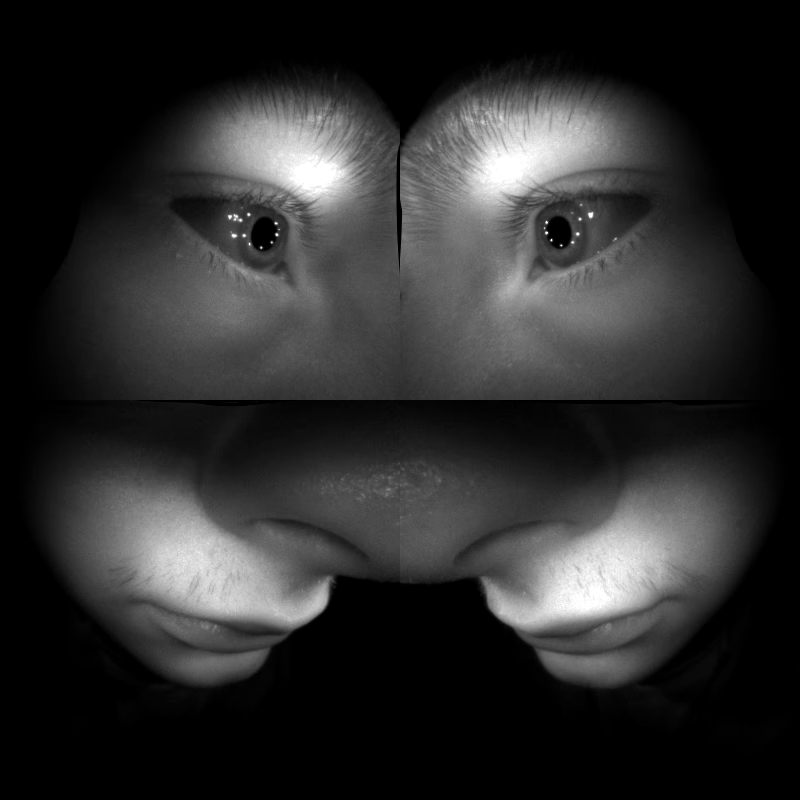}}
    \subcaptionbox*{\scriptsize Config \#1}{\includegraphics[width=0.158\linewidth]{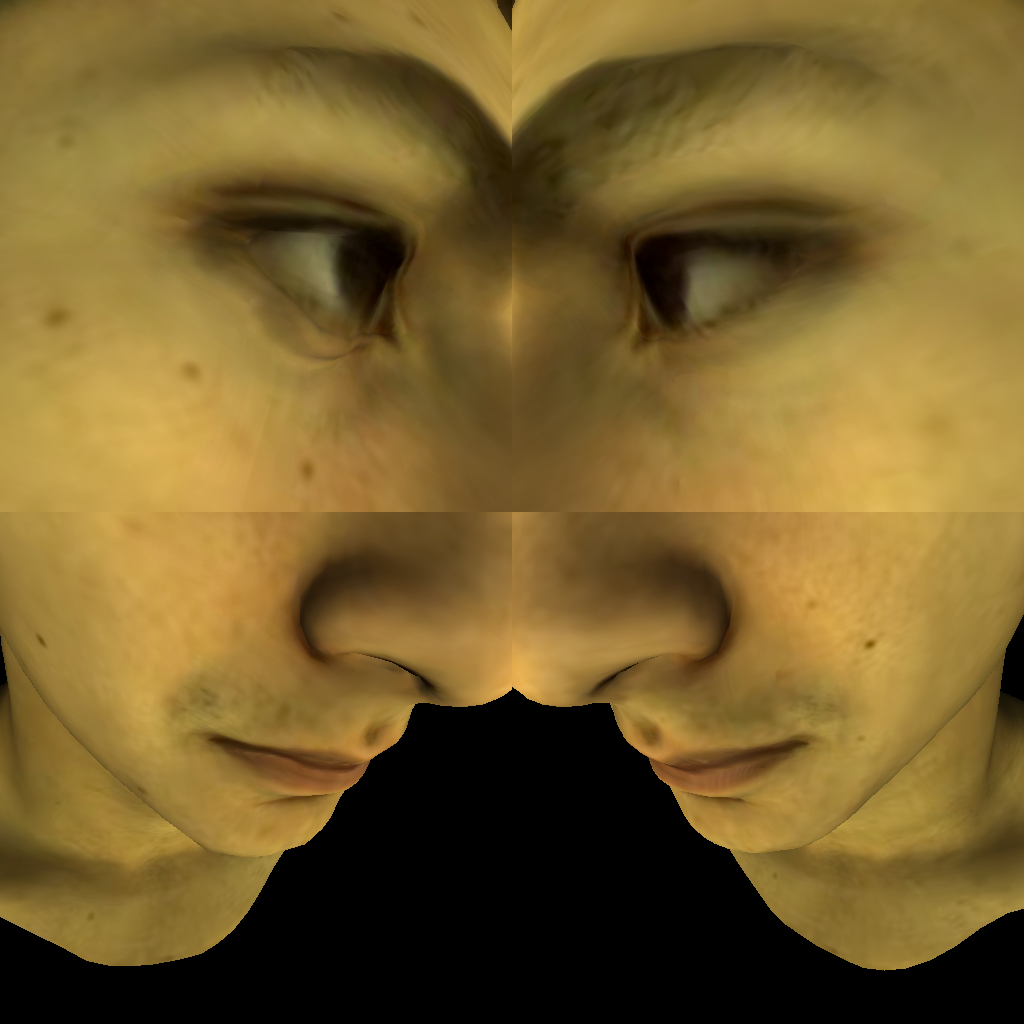}}
    \subcaptionbox*{\scriptsize Config \#2}{\includegraphics[width=0.158\linewidth]{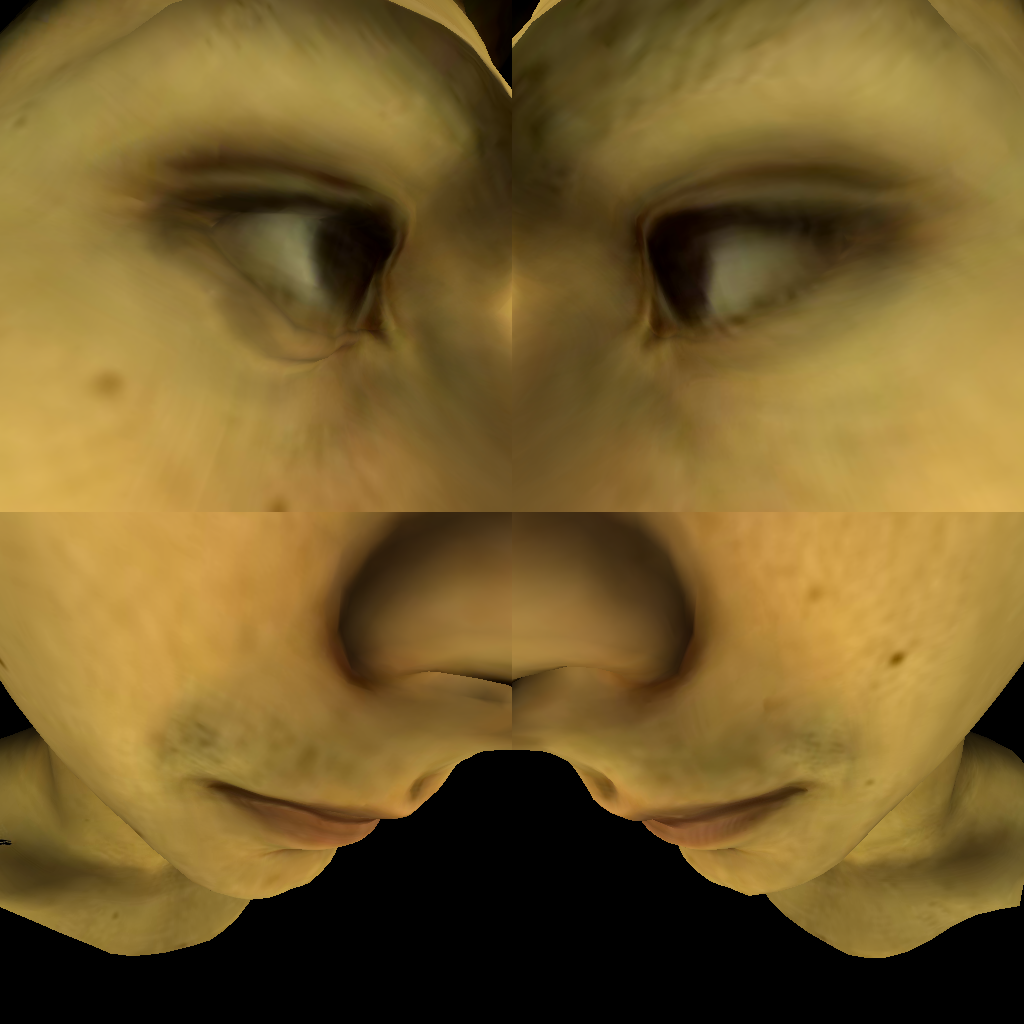}}
    \subcaptionbox*{\scriptsize Config \#3}{\includegraphics[width=0.158\linewidth]{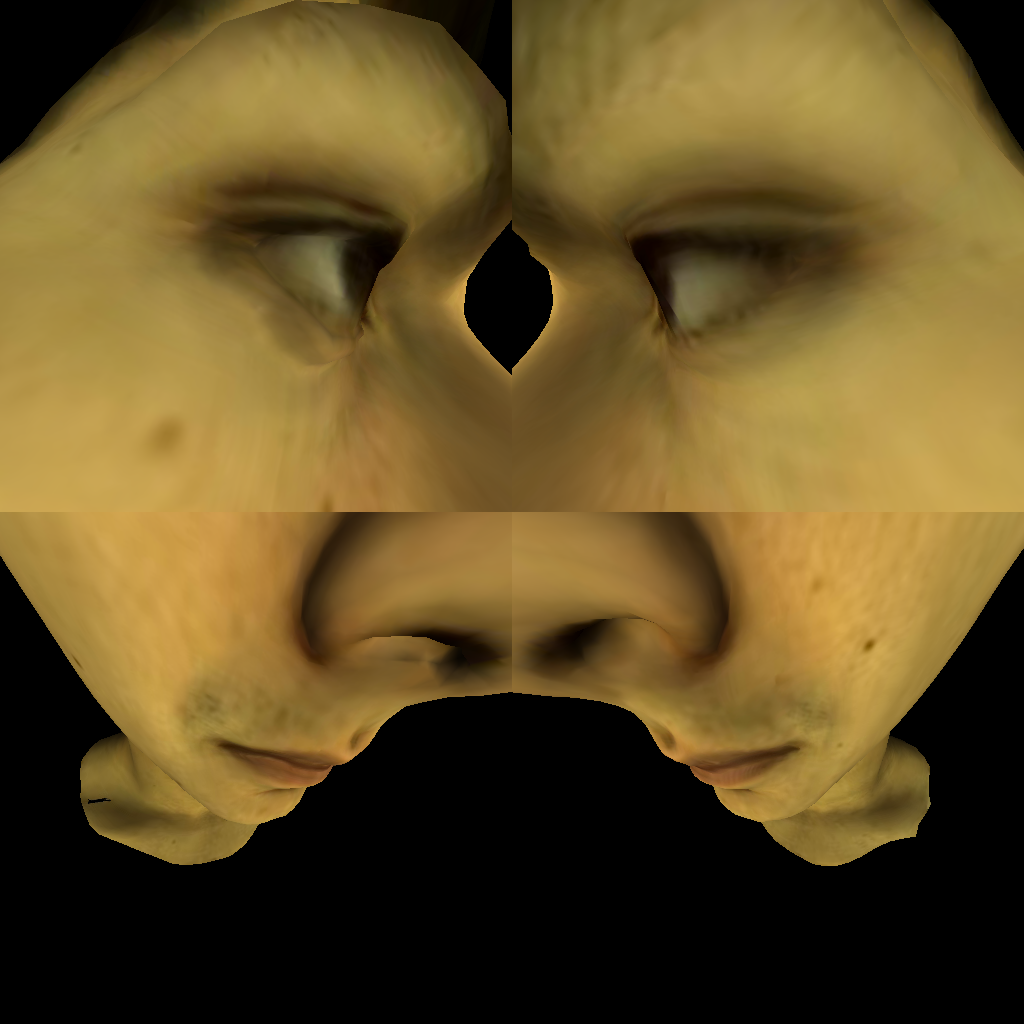}}
    \subcaptionbox*{\scriptsize Config \#4}{\includegraphics[width=0.158\linewidth]{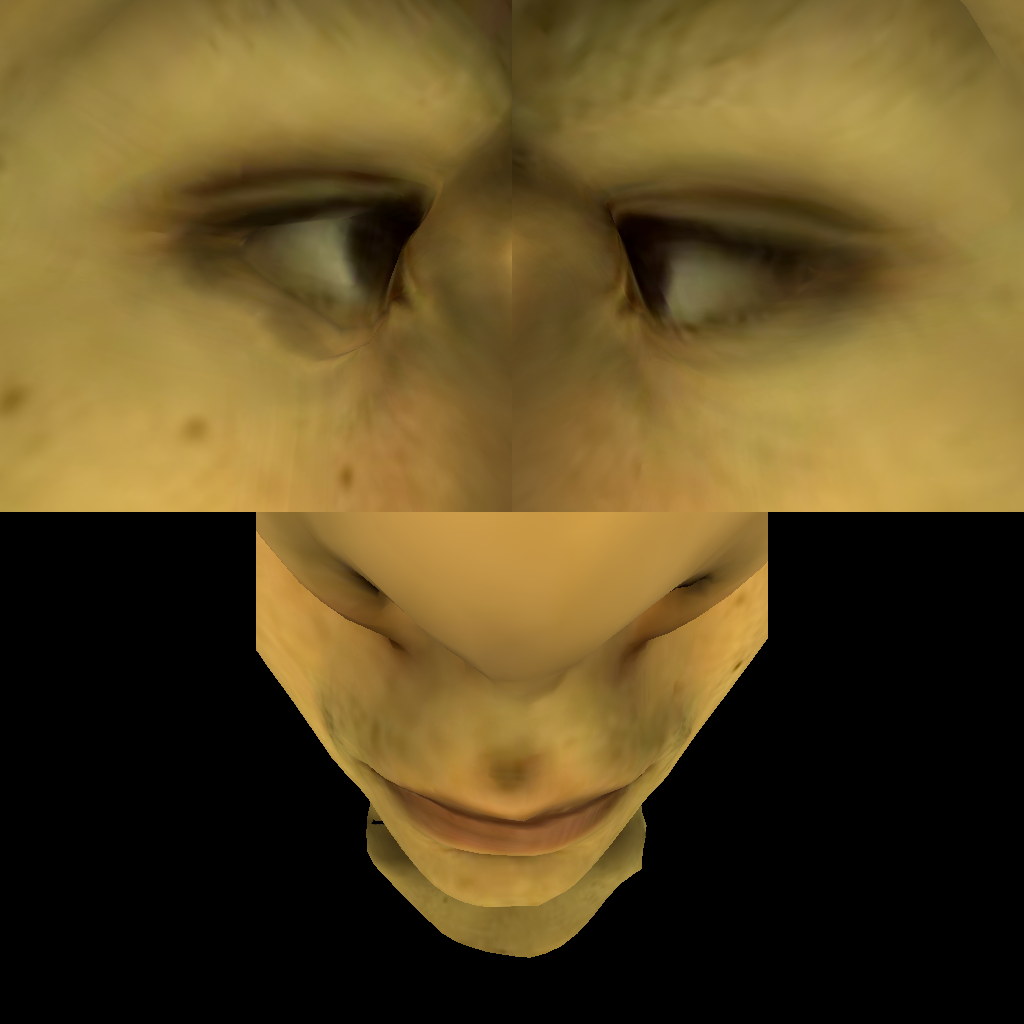}}
    \subcaptionbox*{\scriptsize Config \#5}{\includegraphics[width=0.158\linewidth]{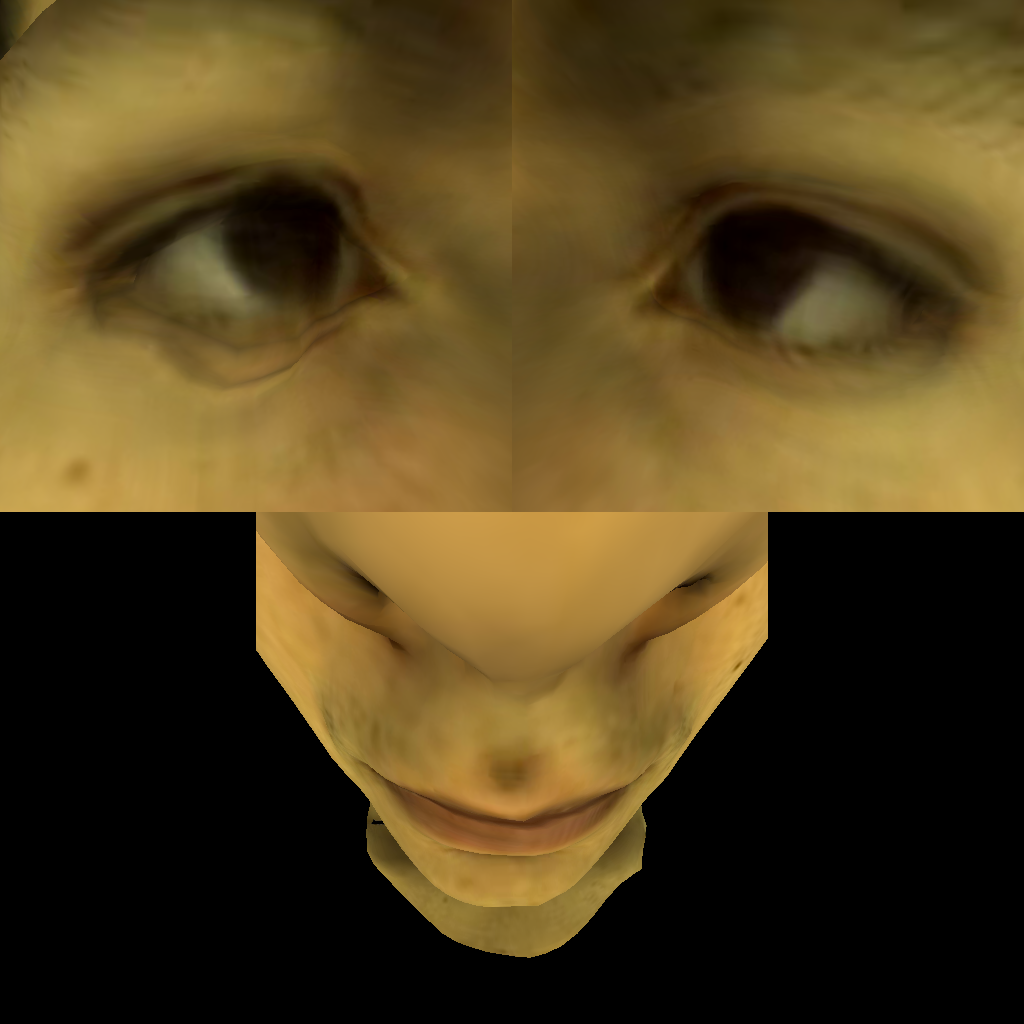}}

    \vspace{-8pt}
    \caption{Camera configurations used in the experiments.}
    \label{fig:cam_cfg_demo}
\end{figure}

\subsection{Datasets}

\begin{table*}[t]
    \centering
    \small
    \begin{tabular}{ccccc}
        \toprule 
        \multirow{2}{*}{Training Dataset} & \multirow{2}{*}{FER Model} & \multirow{2}{*}{\# Params} & Ava256-Scan~\cite{martinezcodec} & 3DRFE~\cite{stratou2011effect} \\
        \cmidrule(lr){4-5} & & & UAR / WAR & Accuracy \\

        \midrule
        \multirow{2}{*}{RAF-DB~\cite{li2017reliable}} 
        & EfficientFace~\cite{zhao2021robust} & 1.3 M & 29.2 ± 3.8 / 31.7 ± 4.2 & 28.2 ± 3.1   \\
        & APViT~\cite{xue2022vision} & 65.2 M & 25.4 ± 2.5 / 29.1 ± 2.7 & 24.2 ± 6.6  \\
        
        \midrule
        \multirow{5}{*}{AffectNet-280K~\cite{mollahosseini2017affectnet}}
         
        & EfficientFace~\cite{zhao2021robust}& 1.3 M & 31.0 ± 4.6 / 33.0 ± 4.4 & 25.0 ± 8.4   \\
        & R-18~\cite{he2016deep} & 11.7 M & 30.6 ± 4.0 / 33.1 ± 4.4 & 28.8 ± 4.0 \\
        & R-50~\cite{he2016deep} & 25.6 M & 30.7 ± 3.2 / 33.1 ± 3.5 & 29.0 ± 5.1 \\
        & Swin-T~\cite{liu2021swin} & 28.3 M & 36.7 ± 5.2 / 38.9 ± 5.5 & 36.2 ± 4.1  \\
        & Swin-B~\cite{liu2021swin} & 87.8 M & 36.3 ± 4.9 / 39.2 ± 5.3 & 36.0 ± 7.8 \\

        \midrule
        \multirow{2}{*}{AffectNet-7K~\cite{mollahosseini2017affectnet}} & R-18~\cite{he2016deep} & 11.7 M & 21.8 ± 5.3 / 19.4 ± 6.5 & 22.4 ± 3.5 \\
        & Swin-T~\cite{liu2021swin} & 28.3 M & 30.4 ± 4.5 / 31.2 ± 5.4 & 23.6 ± 7.1 \\

        \midrule
        \multirow{2}{*}{SynHMC-7K (Ours)} & R-18~\cite{he2016deep} & 11.7 M & \underline{43.1} ± 1.8 / \underline{46.0} ± 2.0 &  \underline{44.5} ± 2.8  \\
        & Swin-T~\cite{liu2021swin} & 28.3 M & \textbf{45.0} ± 2.0 / \textbf{47.9} ± 2.2 & \textbf{48.2} ± 1.9 \\
        
        \bottomrule
    \end{tabular}

    \vspace{-6pt}
    \caption{
    \label{tab:main_ava256_mesh_direct_test} Evaluation results on the \textbf{Ava256-Scan} and \textbf{3DRFE} datasets. 280K and 7K indicate the number of images used for training. The results are averaged over five HMC configurations for Ava256-Scan and three for 3DRFE.
    }
\end{table*}

\textbf{Training Datasets:}
\begingroup 
\renewcommand{\labelitemi}{-}
\begin{itemize}

\item \textbf{Frontal-view (FV) FER datasets.} These datasets contain nearly frontal face portrait images with expression labels.
In particular, we use \textbf{RAF-DB}~\cite{li2017reliable}, a real-world expression dataset with 30K facial images labeled by 40 trained annotators, and \textbf{AffectNet}~\cite{mollahosseini2017affectnet}, a large-scale and diverse dataset for facial expression analysis. We use the subset of AffectNet containing 280K images with human-annotated labels, referred to as \texttt{AffectNet-280K}.

\item \textbf{Synthetic HMC FER dataset.}
We synthesize a dataset of HMC-view images using the proposed SynHMC framework. To address class imbalance, we randomly select a balanced subset of 7K source images (1K per class) from the 280K AffectNet dataset, resulting in \texttt{AffectNet-7K}, for HMC generation. We denote the synthetic HMC dataset as \texttt{SynHMC-7K}. 

\item \textbf{Multi-view FER dataset.} We also train FER models for comparison using the multi-view expression dataset \textbf{MEAD}~\cite{wang2020mead}, which comprises portrait images of 60 actors captured from seven different view angles in a controlled environment, depicting eight expressions.

\item \textbf{Facial texture dataset.}
To train our TSAN for texture refinement, we utilize \textbf{FFHQ-UV}~\cite{bai2023ffhq}, a large-scale facial UV texture dataset with over 50K high-quality UV maps under uniform illumination.
\end{itemize}
\endgroup

\noindent \textbf{Evaluation Datasets:}
\begingroup
\renewcommand{\labelitemi}{-}
\begin{itemize}
\item \textbf{3D Face scan datasets.}
To assess model performance across diverse camera configurations, we utilize face meshes with expression labels from 3D scan datasets and \textit{simulate} cameras to render HMC images for evaluation. We conduct experiments on two 3D face scan datasets. \textbf{Ava256-Scan}~\cite{martinezcodec} captures 3D face meshes from 256 subjects across a diverse range of expressions.
\textbf{3DRFE}~\cite{stratou2011effect} contains high-quality 3D face meshes from 23 subjects, each with one basic expression. Additionally, it provides photometric information that enables photorealistic rendering.
All 3D meshes are registered to the FLAME canonical space using the approach from \cite{li2017learning}.

\item \textbf{Real-world HMC dataset.} 
To the best of our knowledge, \textbf{Ava256-HMC}~\cite{martinezcodec} is the only publicly available dataset suitable for HMC FER with \emph{real} camera setups. It contains HMC video clips recorded from a commercial headset, capturing 256 subjects under a wide range of expressions.
We categorize the clips into basic expressions based on the provided annotations.
For image-based FER, we select the middle frame from each expression clip, as it generally captures the peak expression. The resulting dataset comprises 4,160 images across 7 classes. Details are provided in Appendix A3.
\end{itemize}
\endgroup

\begin{figure*}[t]
    \centering
    \begin{tabular}{@{\hspace{0.1mm}}c@{\hspace{0.5mm}}c@{\hspace{1mm}}c@{\hspace{1mm}}c@{\hspace{1mm}}c@{}}
        \subcaptionbox*{Source Image}{\includegraphics[width=0.1535\textwidth]{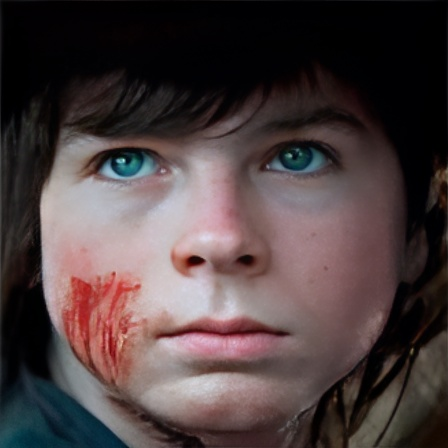}} &
        \subcaptionbox*{SynHMC (Ours)}{\includegraphics[width=0.19\textwidth]{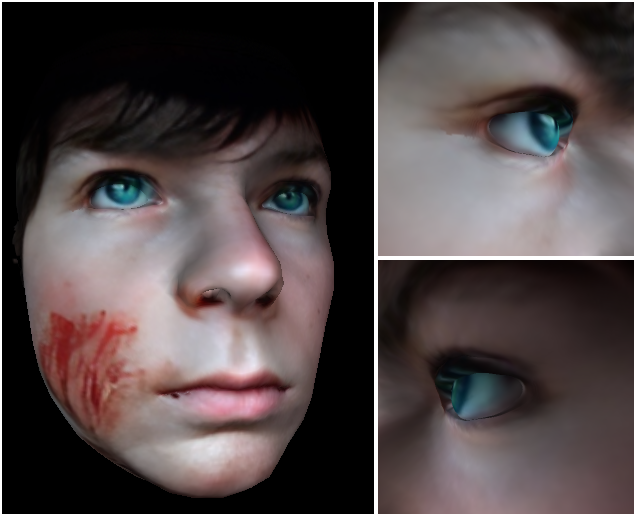}} & \subcaptionbox*{FDS~\cite{chen2019photo}}{\includegraphics[width=0.19\textwidth]{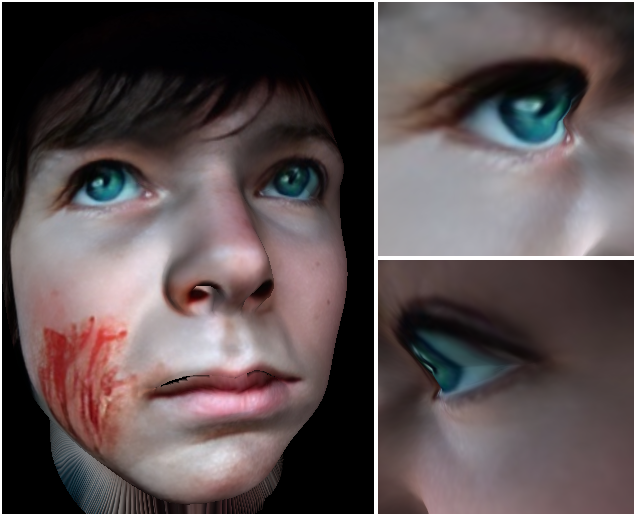}}
        & \subcaptionbox*{FFHQ-UV~\cite{bai2023ffhq}}{\includegraphics[width=0.19\textwidth]{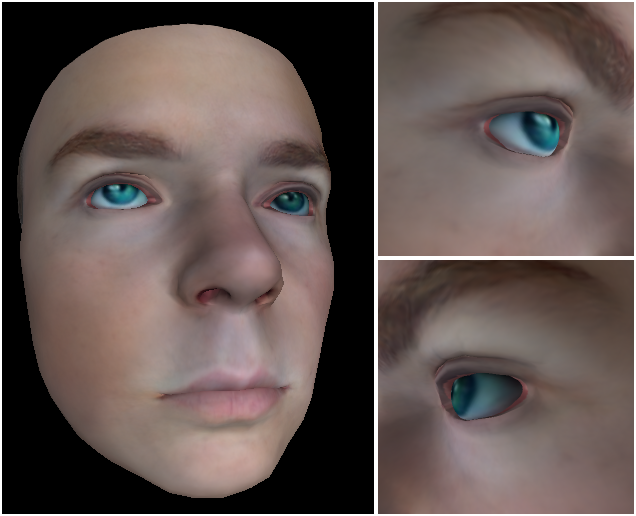}} & \subcaptionbox*{DECA~\cite{feng2021learning}}{\includegraphics[width=0.19\textwidth]{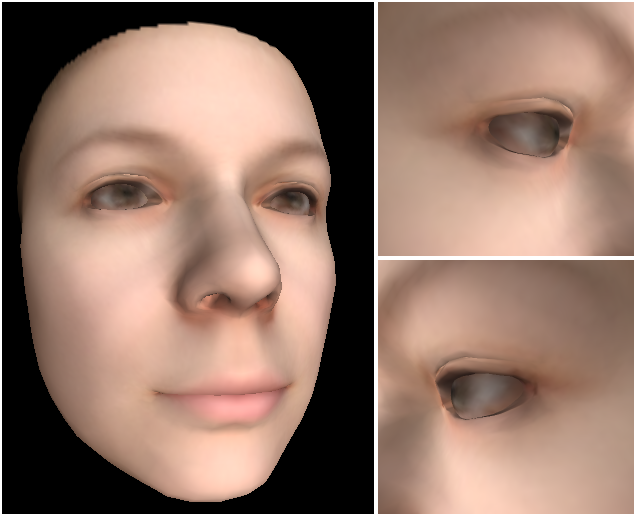}} 
         \\

        \subcaptionbox*{Source Image}{\includegraphics[width=0.1535\textwidth]{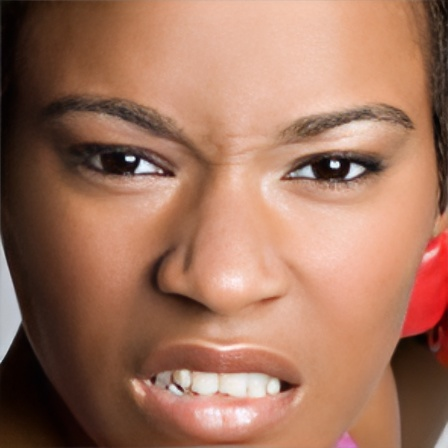}} & \subcaptionbox*{SynHMC (Ours)}{\includegraphics[width=0.19\textwidth]{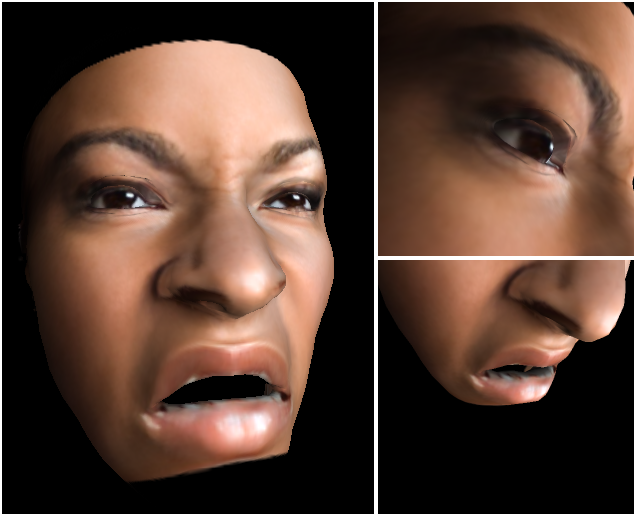}} & \subcaptionbox*{FDS~\cite{chen2019photo}}{\includegraphics[width=0.19\textwidth]{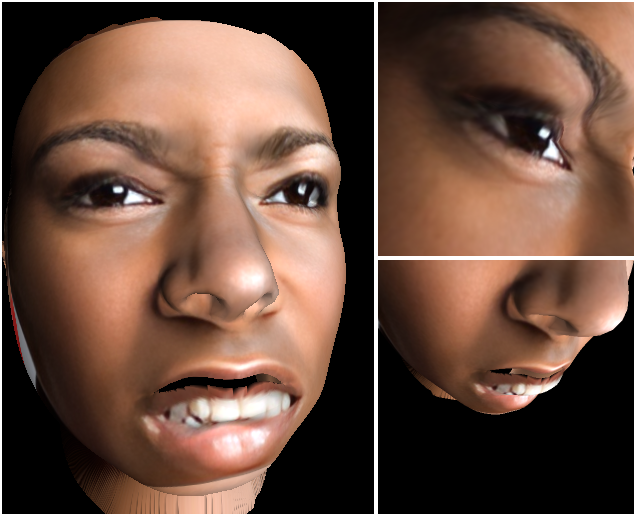}} & \subcaptionbox*{FFHQ-UV~\cite{bai2023ffhq}}{\includegraphics[width=0.19\textwidth]{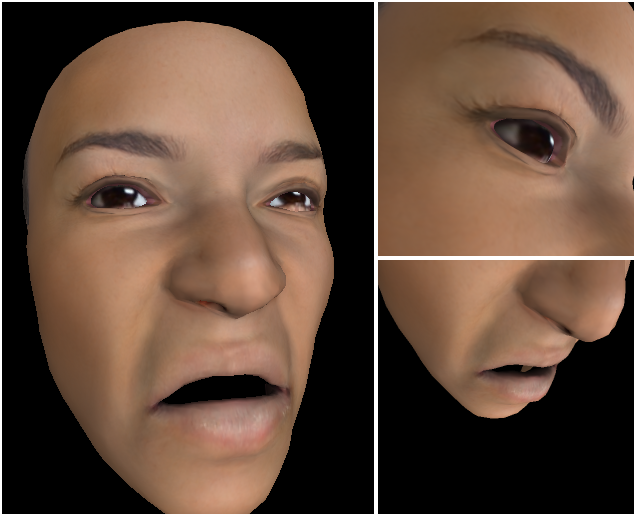}} & \subcaptionbox*{EMOCA~\cite{danvevcek2022emoca}}{\includegraphics[width=0.19\textwidth]{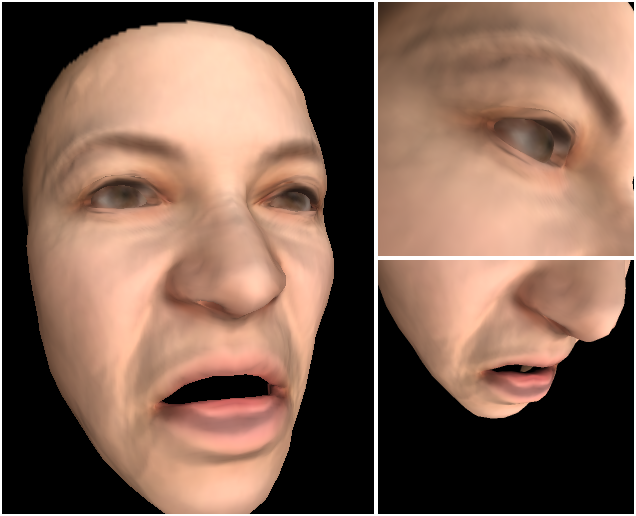}}
         \\
    \end{tabular}

    \vspace{-8pt}
    \caption{Qualitative comparison of 3D face reconstruction methods for HMC image synthesis.}
    \label{fig:cmp_qual_recon}
\end{figure*}

\begin{figure}[t]
    \centering
    \includegraphics[width=0.8\linewidth]{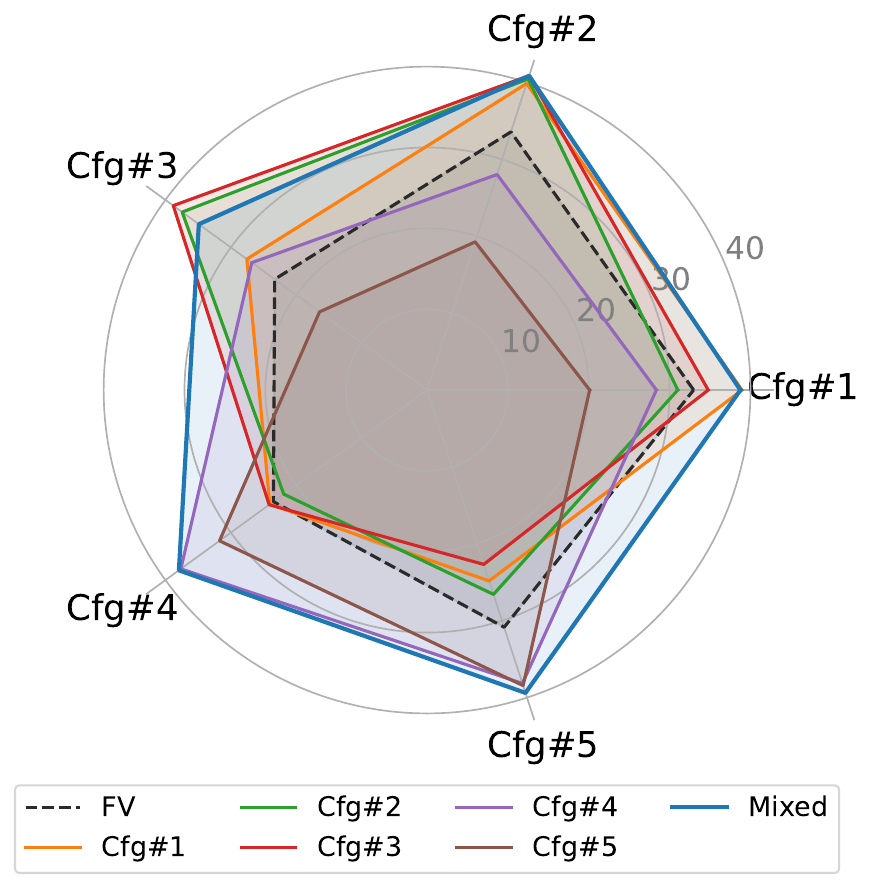}

    \vspace{-5pt}
    \caption{Cross-evaluation of model performance across different HMC cfgs on the \textbf{Ava256-Scan} dataset. Each line represents a model trained using a specific cfg, and each tick corresponds to a test dataset generated from that cfg. UAR is shown in the figure.}
    \label{fig:cross_test_cam_cfgs}
\end{figure}

\vspace{-4pt}
\subsection{Evaluation Protocols}

For evaluation on 3D face scan datasets, we directly test models on HMC images rendered from five different camera configurations. 
As shown in~\cref{fig:cam_cfg_demo}, these configurations are selected based on existing headset solutions~\cite{wei2019vr,jourabloo2022robust,martinezcodec}. Cameras within each configuration share the same FoV. The FoV increases progressively from Config \#1 to \#3 (so as distortion), while in Config \#4 and \#5, the number of views is reduced to three.
We test on each configuration and report the averaged results. 

For real HMC FER evaluation, we conduct direct testing and few-shot evaluation using 4-fold cross-validation approach.
Specifically, we initialize models with pre-trained weights, \texttt{fine-tune} on training fold, and evaluate on the test fold. Additionally, we adopt \texttt{K-NN}~\cite{caron2021emerging} and \texttt{linear probe}~\cite{chen2020simple,caron2021emerging,he2022masked} to evaluate the quality of the learned feature extractors.

\noindent\textbf{Metrics.}
Following previous work~\cite{xue2022vision, wang2023rethinking, zhang2024generalizable, zhao2025enhancing} on FER, we report the average test accuracy across all classes for class-balanced dataset (\ie, 3DRFE). For imbalanced dataset (\ie, Ava256), we report both unweighted average recall (UAR) and weighted average recall (WAR). UAR averages recall equally across classes, while WAR weights recall by class proportion. Unless stated otherwise, UAR and WAR values are presented as percentages.

\vspace{-3pt}
\subsection{Implementation Details}
We implement the data synthesis framework and model training using PyTorch~\cite{ansel2024pytorch} and PyTorch3D~\cite{ravi2020accelerating}.
For training FER models, we use the Adam optimizer to minimize cross-entropy loss.
Cosine annealing scheme~\cite{loshchilov2016sgdr} is used with initial learning rate $1\times10^{-4}$. All models are trained for 10 epochs, with a mini-batch size of 32. During training, on-the-fly data augmentation is applied, including 3D augmentations that randomly perturb mesh and HMC positions, as well as 2D augmentations such as random cropping and horizontal flipping.
At test time, images are resized to $256\times256$ pixels and fed into the model. All experiments are conducted on an NVIDIA V100 GPU.

\vspace{-4pt}
\subsection{Results on 3D Face Scan Datasets}

We conduct experiments on face scan datasets to evaluate FER performance across various HMC configurations.

\noindent\textbf{Comparisons of training data sources.} To evaluate the impact of training data sources, we compare models trained on our synthetic HMC images with those trained on alternative datasets, including the two frontal-view FER datasets, RAF-DB and AffectNet. We evaluate and report results across multiple network architectures, including FER-specific models EfficientFace~\cite{zhao2021robust} and APViT~\cite{xue2022vision}, as well as general vision models ResNet~\cite{he2016deep} and Swin Transformer~\cite{liu2021swin}.
As shown in~\cref{tab:main_ava256_mesh_direct_test}, models trained on our synthetic HMC data consistently outperform those trained on existing frontal-view datasets.
Notably, despite using only 7K images for HMC synthesis, models trained on synthetic HMC data outperform larger counterparts trained on 280K frontal images. This highlights the importance of view-specific HMC images for training and demonstrates that our synthetic data effectively reduces the domain gap from FV to HMC FER.

\begin{table}[t]
    \centering
    \small
    \begin{tabular}{lccc}
        \toprule
        \multirow{2}{*}{Method} & \multirow{2}{*}{\makecell{Speed \\ (FPS)}} & Ava256-Scan & 3DRFE  \\
        \cmidrule(lr){3-4} & & UAR / WAR & Acc. \\

        \midrule
        DECA~\cite{feng2021learning} & 22.24 & 26.3 / 26.5 & 32.9 \\
        EMOCA~\cite{danvevcek2022emoca} & 20.08 & 27.2 / 27.7 & 39.3 \\
        FDS~\cite{chen2019photo} & 0.39  & \underline{39.8} / \underline{40.9} & \underline{45.3}  \\
        FFHQ-UV~\cite{bai2023ffhq} & 0.01 & 35.7 / 37.2 & 42.6 \\

        \midrule
        SynHMC (Ours) & 5.03 & \textbf{45.0} / \textbf{47.9} & \textbf{48.2} \\
        
        \bottomrule
    \end{tabular}

    \vspace{-5pt}
    \caption{
        \label{tab:cmp_face_recon}
        Evaluation of different face reconstruction methods for HMC synthesis. AffectNet-7K is used for training data generation, with Swin-T as the classifier backbone. The results are averaged over five HMC configurations.
    }
\end{table}

\begin{table*}[t]
    \centering
    \small
    \begin{threeparttable}
        \begin{tabular}{cccccc}
            \toprule
            \multirow{2}{*}{\makecell{Pre-training Dataset}} & \multirow{2}{*}{\makecell{ Reconstruction  Method}} & Direct Test & 
            K-NN & Linear Probe & Fine-tune \\ 
            \cmidrule(lr){3-6} &  & UAR / WAR & UAR / WAR & UAR / WAR & UAR / WAR \\ 
    
            \midrule
            None & \multirow{4}{*}{None} & -- & -- & -- & 14.56 / 17.82 \\
            ImageNet~\cite{deng2009imagenet} & & -- & -- & -- & 46.40 / 47.96 \\
            MEAD~\cite{wang2020mead} &  & 15.78  / 16.27 & 23.49 / 25.40 & 23.68 / 26.31 & 22.45 / 25.97 \\
            AffectNet-7K~\cite{mollahosseini2017affectnet} & & 19.13 / 18.20 & 33.65 / 34.43 & 31.74 / 34.04 & 50.27 / 51.75 \\

            \midrule
            \multirow{4}{*}{SynHMC-7K}& DECA~\cite{feng2021learning} & 21.35 / 20.02 & 33.73 / 35.42 & 29.78 / 32.84 & 46.82 / 48.25 \\
            & EMOCA~\cite{danvevcek2022emoca} & 19.38 / 18.75 & 33.27 / 34.73 & 32.04 / 35.25 & 49.41 / 50.89 \\
            & FDS~\cite{chen2019photo} & 21.45 / 19.62 & \underline{37.66} / \underline{39.13} & \underline{35.68} / \underline{38.42} & 50.73 / 52.05 \\
            & FFHQ-UV~\cite{bai2023ffhq} & \underline{25.78} / \underline{26.49} & 36.58 / 38.33 & 34.23 / 37.33 & \underline{51.76} / \underline{52.90} \\

            \midrule
            SynHMC-7K & SynHMC (Ours)  & \textbf{27.93} / \textbf{27.79} & \textbf{40.53} / \textbf{42.03} & \textbf{38.27} / \textbf{41.04} & \textbf{53.42} / \textbf{54.40} \\

            \bottomrule
        \end{tabular}
    \end{threeparttable}

    \vspace{-7pt}
    \caption{\label{tab:res_ava_hmc}Performance on the \textbf{Ava265-HMC} dataset with Swin-T backbone.  \emph{Config \#1} is used to synthesize training HMC. In pre-training dataset, "None" refers to randomly initialized FER models, while in reconstruction method, it denotes training with frontal-view images.}
\end{table*}

\noindent\textbf{Comparisons of HMC configurations.}
To understand the generalization capability across different HMC settings, 
we perform cross-validation on data from five HMC configurations. Specifically, we evaluate models trained on different sources by testing them on data generated from each HMC configuration separately. As shown in~\cref{fig:cross_test_cam_cfgs}, models perform well when trained and tested on the same configuration but fail to generalize across different HMC configurations due to significant differences. A key advantage of data synthesis is the ability to generate and combine data from various HMC views for training. As can be seen, the \texttt{Mixed} model, trained on data from all five views, achieves significantly improved generalization across configurations.

\begin{table}[t]
    \centering
    \small
    \begin{tabular}{lc}
        \toprule
        Method & UAR / WAR  \\
        \midrule
        Baseline (w/o refinement) & 39.5 ± 1.0 / 42.1 ± 0.7 \\
        w/ ARAP only & 44.2 ± 2.4 / 46.0 ± 2.0 \\
        w/ TSAN only & 42.0 ± 1.9 / 44.5 ± 2.2 \\
        w/ ARAP and TSAN & 45.0 ± 2.0 / 47.9 ± 2.2 \\
        \bottomrule
    \end{tabular}

    \vspace{-7pt}
    \caption{FER performance on the \textbf{Ava256-Scan} dataset with different texture refinement options.}
    \label{tab:refinement}
\end{table}

\noindent\textbf{Comparisons of face reconstruction methods.}
We compare our approach with state-of-the-art face reconstruction methods, including DECA~\cite{feng2021learning}, EMOCA~\cite{danvevcek2022emoca}, FDS~\cite{chen2019photo}, and FFHQ-UV~\cite{bai2023ffhq}. Qualitative comparisons of the generated HMC images are provided in~\cref{fig:cmp_qual_recon}. As observed, methods using linear and GAN-based texture models fail to capture appearance details from the source image. While FDS unwraps textures from images, misalignment results in noticeable artifacts, such as mismatches between shape and texture (\eg, teeth appearing despite not being modeled in shape). Our method reduces these artifacts and generates HMC images with detailed expressions and diverse appearances.
\Cref{tab:cmp_face_recon} provides FER results using HMC training data generated from these methods.
As seen, models trained with detailed expressions achieve better performance in HMC FER.

\begin{figure}[t]
    \centering
    \includegraphics[width=0.3\linewidth]{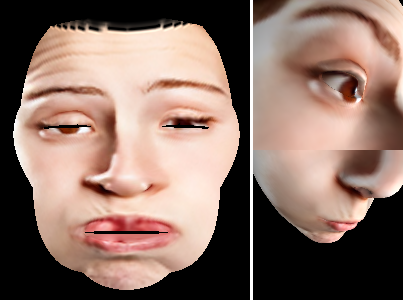} 
    \includegraphics[width=0.3\linewidth]{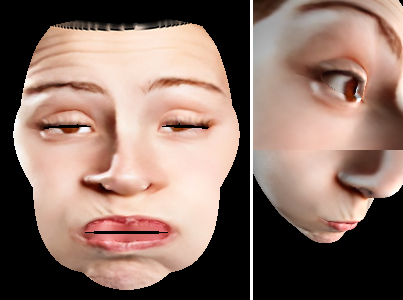} \includegraphics[width=0.3\linewidth]{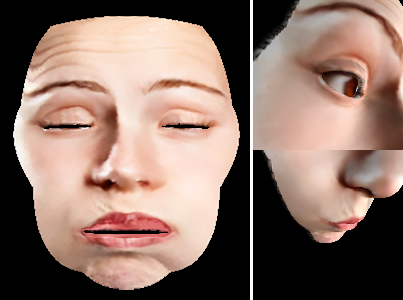}
    
    \vspace{2pt}
    \subcaptionbox*{w/o Refinement}{\includegraphics[width=0.30\linewidth]{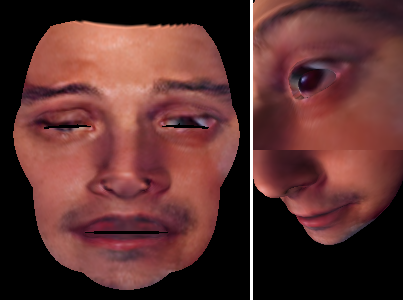}}
    \subcaptionbox*{w/ ARAP}{\includegraphics[width=0.30\linewidth]{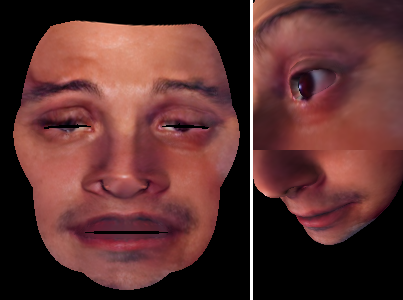}}
    \subcaptionbox*{w/ ARAP \& TSAN}{\includegraphics[width=0.30\linewidth]{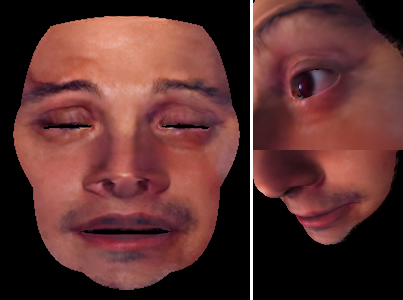}}

    \vspace{-7pt}
    \caption{Comparisons of different texture refinement methods.}
    \label{fig:refinement}
\end{figure}

\noindent\textbf{Impact of texture refinement.}
We begin with a baseline model using EMOCA-estimated camera parameters without texture alignment for HMC data generation, then incorporate ARAP, TSAN, and their combination.
\Cref{fig:refinement} provides a visual comparison.
As shown, ARAP improves alignment using off-the-shelf dense landmarks but still struggles in regions with inaccurate landmarks, such as the eyes and nose. TSAN further improves alignment by leveraging the estimated texture-space rectified flow.
As shown in~\Cref{tab:refinement}, better alignment also enhances FER performance by reducing artifacts, enabling models to focus on expression-related features rather than noise.

\vspace{-3pt}
\subsection{Results on Real-world HMC dataset}

We further evaluate model performance in \emph{real-world scenarios} using the Ava256-HMC dataset. Results are shown in \cref{tab:res_ava_hmc}.

As observed, all models struggles in direct test due to the sim-to-real gap (\eg, headset donning, color intensity, lighting conditions). However, K-NN and linear probing results show that our feature extractors effectively capture expression features despite being trained on synthetic data, achieving better performance over models trained on FV FER datasets.
For fine-tuning, the model trained from scratch fails to converge due to the limited amount of training data. Similarly, the MEAD-trained model performs poorly as the dataset contains fewer than 60 subjects. This highlights the importance of scale and diversity in training data.
Again, the model trained on our synthetic data outperform other approaches.  Notably, the performance gap narrows in Ava256-HMC, likely due to Ava256’s relatively simple expressions and uniform scripts, allowing models to generalize well with minimal fine-tuning data.

\section{Conclusion and Limitations}

\vspace{-5pt}
We investigate the domain gap between frontal-view and HMC FER, and propose SynHMC, a data synthesis framework that transforms existing frontal-view datasets into HMC datasets to address data scarcity.

While effective, our method has several limitations. First, it primarily addresses the camera perspective gap between FV and HMC images, which we consider most critical for FER. However, other domain factors such as lighting and color differences are not explicitly handled. We plan to incorporate image relighting and editing techniques in future work to address these aspects. Second, our method inherits the limitations of FLAME, which, like many 3DMMs, does not model the inner-mouth region. Incorporating a more expressive 3DMM could further improve reconstruction quality and overall performance.
{
    \small
    \bibliographystyle{ieeenat_fullname}
    \bibliography{main}
}

\clearpage

\appendix
\renewcommand{\thesection}{A\arabic{section}}
\renewcommand{\thefigure}{A\arabic{figure}}
\renewcommand{\thetable}{A\arabic{table}}
\renewcommand{\theequation}{A\arabic{equation}}
\setcounter{page}{1}
\setcounter{section}{0}
\setcounter{figure}{0}
\setcounter{table}{0}
\setcounter{equation}{0}
\maketitlesupplementary




\section{Texture-Space Alignment Network}
\label{supp sec: tsan}

\begin{figure}[t]
    \centering
    \subcaptionbox*{\scriptsize Anchor}{\includegraphics[width=0.24\linewidth]{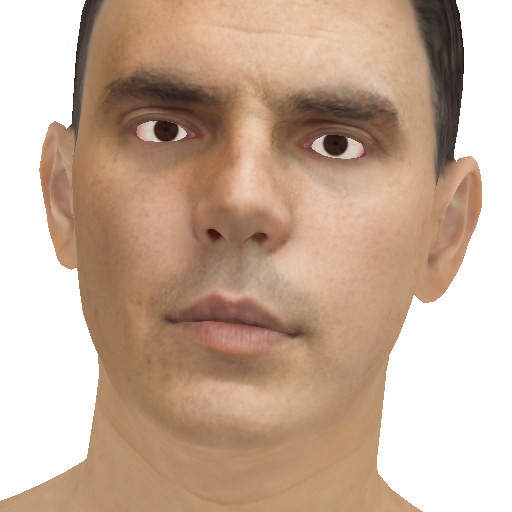}}
    \subcaptionbox*{\scriptsize Perturbed Camera}{\includegraphics[width=0.24\linewidth]{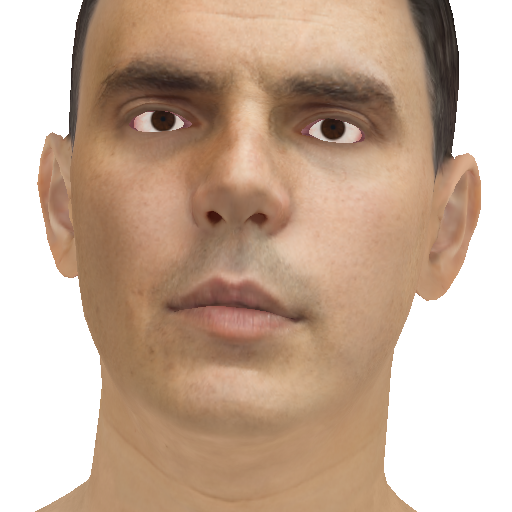}}
    \subcaptionbox*{\scriptsize Perturbed Mesh}{\includegraphics[width=0.24\linewidth]{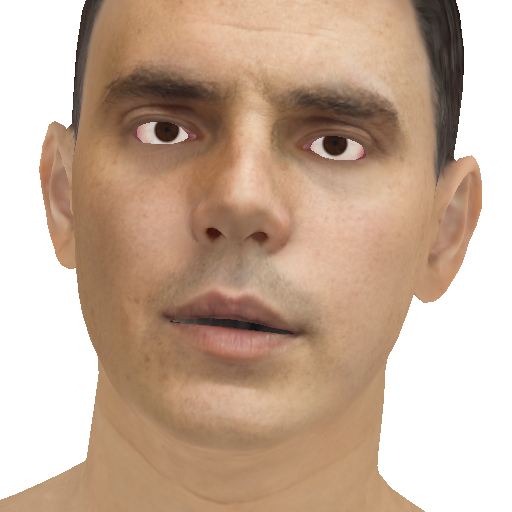}}
    \subcaptionbox*{\scriptsize Mesh\&Cam}{\includegraphics[width=0.24\linewidth]{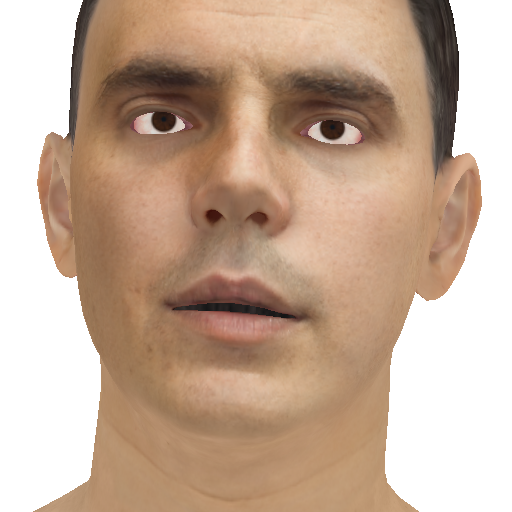}}

    \subcaptionbox*{UV Template}{\includegraphics[width=0.24\linewidth]{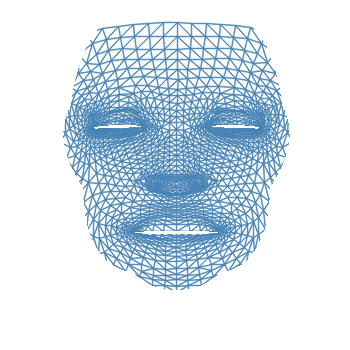}}
    \subcaptionbox*{\scriptsize Difference}{\includegraphics[width=0.24\linewidth]{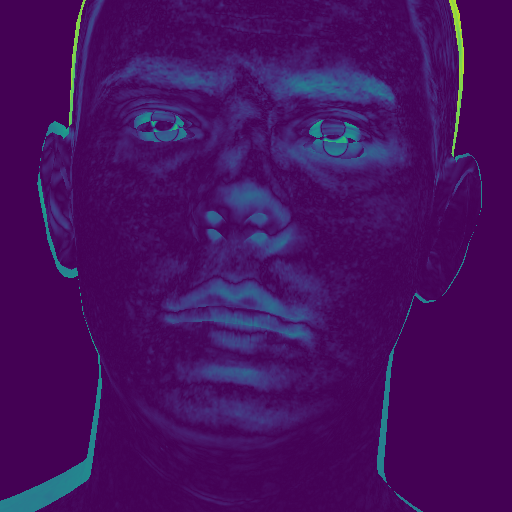}}
    \subcaptionbox*{\scriptsize Difference}{\includegraphics[width=0.24\linewidth]{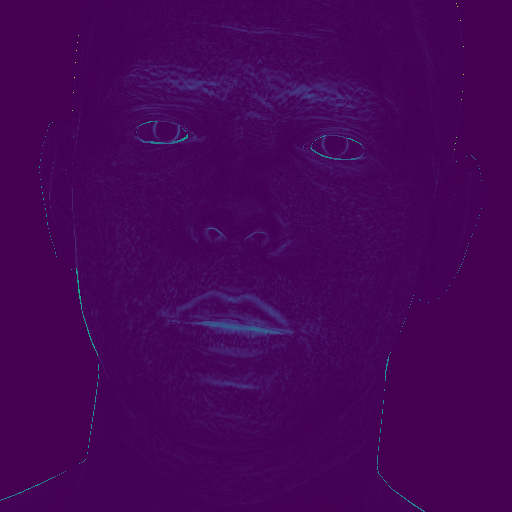}}
    \subcaptionbox*{\scriptsize Difference}{\includegraphics[width=0.24\linewidth]{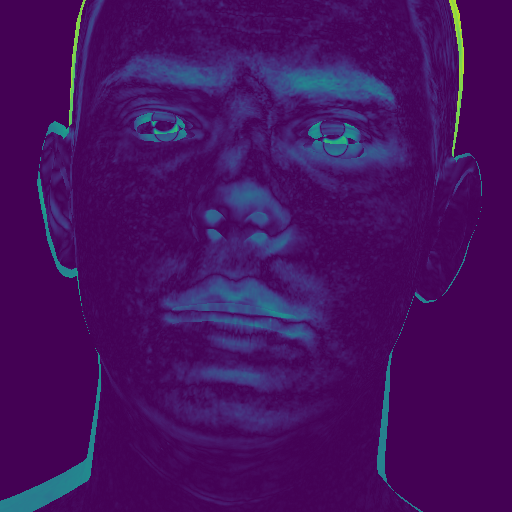}}

    \subcaptionbox*{\scriptsize GT Texture}{\includegraphics[width=0.24\linewidth]{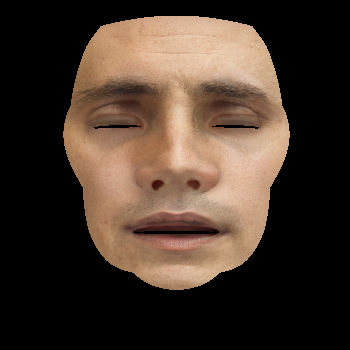}}
    \subcaptionbox*{\scriptsize LQ Texture1}{\includegraphics[width=0.24\linewidth]{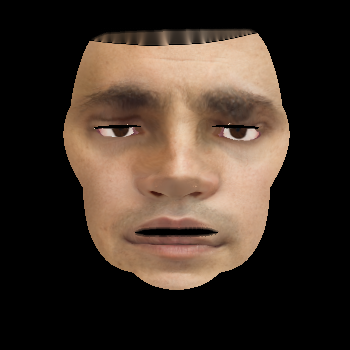}}
    \subcaptionbox*{\scriptsize LQ Texture2}{\includegraphics[width=0.24\linewidth]{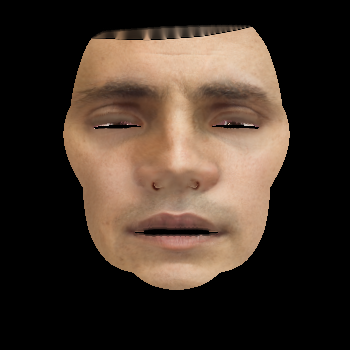}}
    \subcaptionbox*{\scriptsize LQ Texture3}{\includegraphics[width=0.24\linewidth]{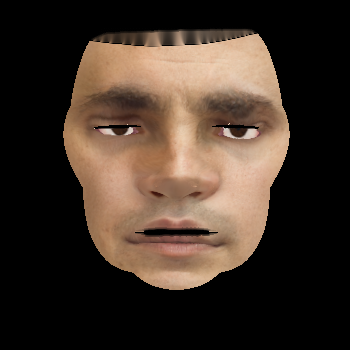}}

    \vspace{-5pt}
    \caption{Illustration of synthetic texture pairs for TSAN training. Columns: anchor, perturbed mesh, perturbed camera, and both. Rows: reprojected image, difference from anchor, and unwrapped texture maps. LQ texture maps are fed into TSAN for rectified flow estimation.}
    \label{fig:synth_pair_tsan}
\end{figure}

\paragraph{Data synthesis for TSAN training.} \Cref{alg:syn_tsan} outlines the algorithm for generating a single pair of ground-truth and misaligned textures for TSAN training. During training, we randomly select EMOCA-estimated face meshes from AffectNet and different target texture maps from the FFHQ-UV dataset and use them to generate diverse pairs for TSAN training. \Cref{fig:synth_pair_tsan} presents additional synthesized data pairs with individual perturbations in shape code, expression code, and camera parameters.

\begin{algorithm}
\caption{Generating a data pair for TSAN training}\label{alg:syn_tsan}
\begin{algorithmic}[1]
\State \textbf{Input:} FLAME params $\boldsymbol{\beta}, \boldsymbol{\psi}, \boldsymbol{\theta}$, camera params $\boldsymbol{c}$, target texture map $\mathbf{I}_\mathrm{tex}^\mathrm{gt}$
\State \textbf{Output:} Misaligned texture $\mathbf{I}_\mathrm{tex}^\mathrm{lq}$

\State \green{// Render Reference Image}

\State Build anchor mesh $(\mathbf{V}, \mathbf{F}) \gets M(\boldsymbol{\beta}, \boldsymbol{\psi}, \boldsymbol{\theta})$
\State Render reprojected image $\mathbf{I}_\mathrm{ref} \gets \mathcal{R}(\mathbf{V}, \mathbf{F}, \mathbf{I}_\mathrm{tex}^\mathrm{gt}, \boldsymbol{c})$

\State \green{// Perturb Mesh}

\State Sample a scalar \( s_{shape} \sim \mathcal{N}(0, 0.3^2) \) and perturb the shape parameters $\boldsymbol{\beta}' \gets (1 + s_{shape}) \cdot \boldsymbol{\beta}$

\State Sample a scalar \( s_{exp} \sim \mathcal{N}(0, 0.3^2) \) and perturb the expression parameters $\boldsymbol{\psi}' \gets (1 + s_{exp}) \cdot \boldsymbol{\psi}$

\State Build perturbed mesh $(\mathbf{V}', \mathbf{F}) \gets M(\boldsymbol{\beta}', \boldsymbol{\psi}', \boldsymbol{\theta})$

\State \green{// Perturb Camera}
\State Convert camera orientation to Euler angles (in degree) $\boldsymbol{R} \in \mathbb{R}^3$, sample a vector $\boldsymbol{\delta}_R \in \mathbb{R}^3 \sim \mathcal{N}(0, 3^2)$ and perturb $\boldsymbol{R}' \gets \boldsymbol{R} + \boldsymbol{\delta}_R$
\State Sample a vector $\boldsymbol{\delta}_t \in \mathbb{R}^3 \sim \mathcal{N}(0, 0.001^2)$ and perturb camera translation $\boldsymbol{t}' \gets \boldsymbol{t} + \boldsymbol{\delta}_T$
\State Combine $\boldsymbol{R}'$ and $\boldsymbol{t}'$ to get new camera parameters $\boldsymbol{c}'$

\State \green{// Unwrap the reference image using the perturbed mesh and camera}
\State Unwrap to get misaligned texture map  $\mathbf{I}_\mathrm{tex}^\mathrm{lq} \gets \mathcal{I}(\mathbf{I}_\mathrm{ref}, \mathbf{V}', \mathbf{F}, \boldsymbol{c}')$

\State \textbf{Return} $\mathbf{I}_\mathrm{tex}^\mathrm{lq}$
\end{algorithmic}
\end{algorithm}

\paragraph{Network architecture.} We use a simple U-Net for TSAN to estimate the rectified flow, which takes as input a texture map and produces a 2D flow field as output. We also adopt a learnable positional encoding map to encode pixel locations and concatenate it with the outputs after the first convolutional layer. \Cref{tab:tsan_arch} presents the detailed architecture of TSAN.

\paragraph{Training details.}
We employ the Adam optimizer to minimize the L1 loss between the misaligned and ground-truth texture maps. A cosine annealing schedule is applied with an initial learning rate of $3\times10^{-4}$. All models are trained for 15 epochs with a mini-batch size of 4.

\begin{figure*}[t]
    \centering
    \includegraphics[width=0.17\linewidth]{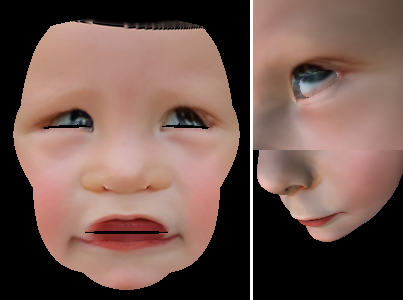}
    \includegraphics[width=0.17\linewidth]{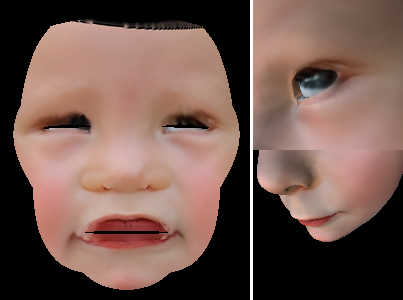}
    \includegraphics[width=0.11\linewidth]{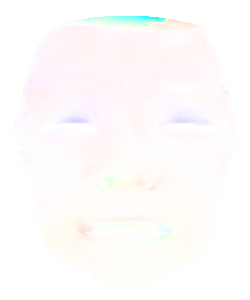}
    \includegraphics[width=0.17\linewidth]{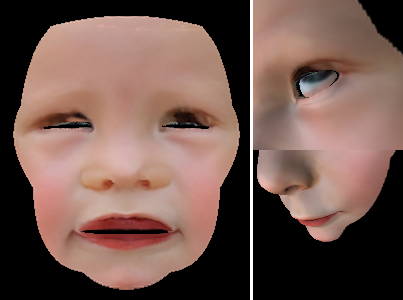}
    \includegraphics[width=0.11\linewidth]{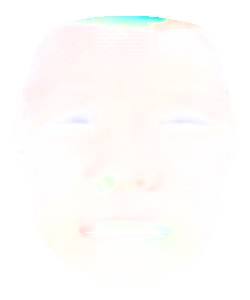}
    \includegraphics[width=0.17\linewidth]{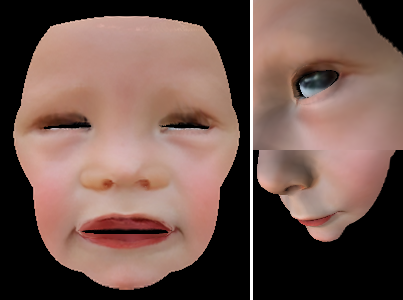}

    \includegraphics[width=0.17\linewidth]{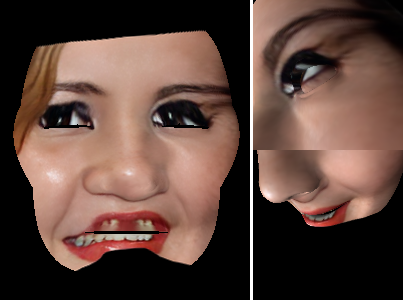}
    \includegraphics[width=0.17\linewidth]{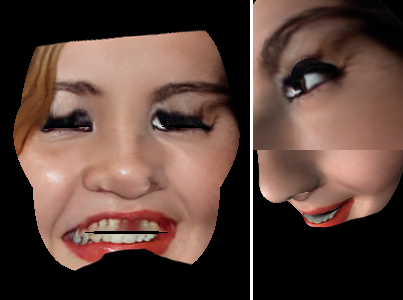}
    \includegraphics[width=0.11\linewidth]{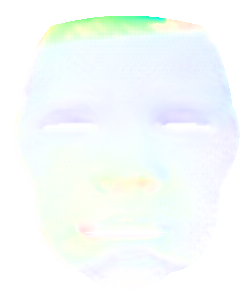}
    \includegraphics[width=0.17\linewidth]{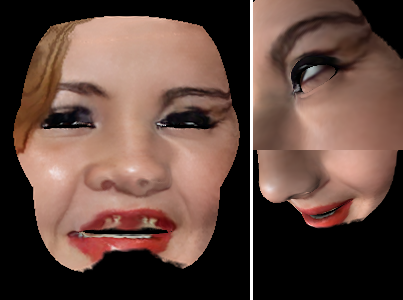}
    \includegraphics[width=0.11\linewidth]{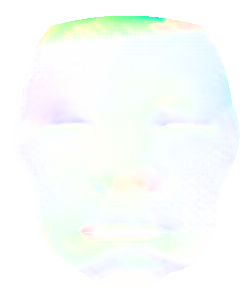}
    \includegraphics[width=0.17\linewidth]{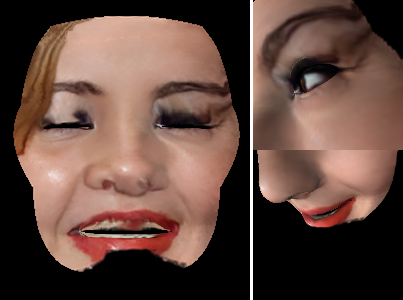}

    \includegraphics[width=0.17\linewidth]{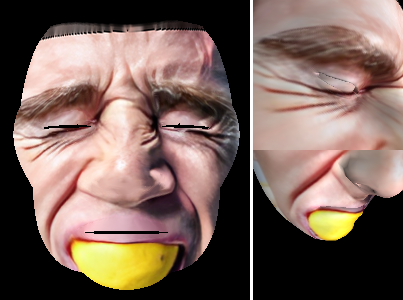}
    \includegraphics[width=0.17\linewidth]{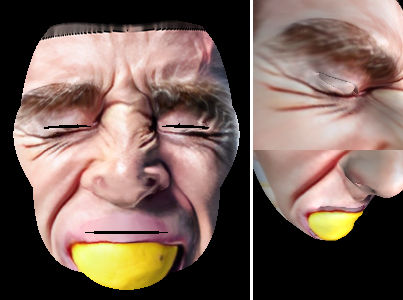}
    \includegraphics[width=0.11\linewidth]{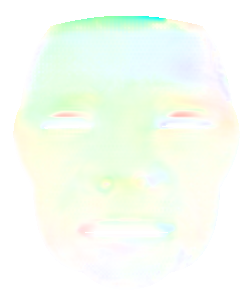}
    \includegraphics[width=0.17\linewidth]{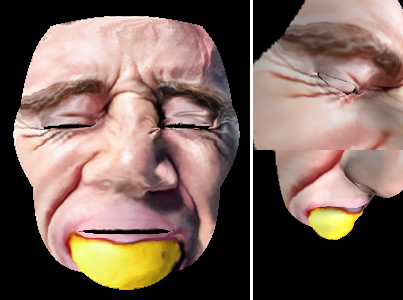}
    \includegraphics[width=0.11\linewidth]{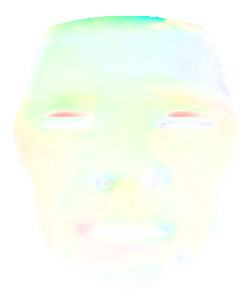}
    \includegraphics[width=0.17\linewidth]{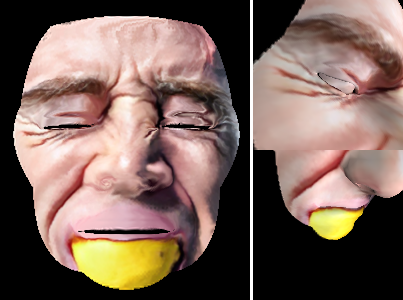}
    
    \subcaptionbox*{(a)}{\includegraphics[width=0.17\linewidth]{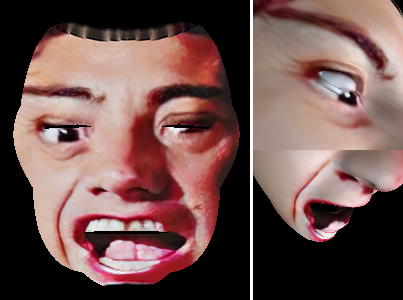}}
    \subcaptionbox*{(b)}{\includegraphics[width=0.17\linewidth]{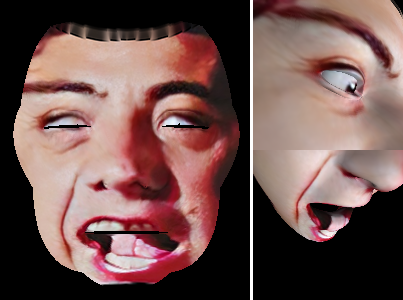}}
    \subcaptionbox*{(c)}{\includegraphics[width=0.11\linewidth]{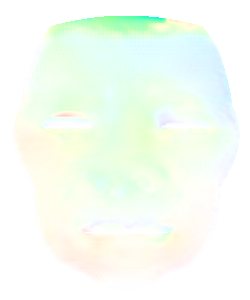}}
    \subcaptionbox*{(d)}{\includegraphics[width=0.17\linewidth]{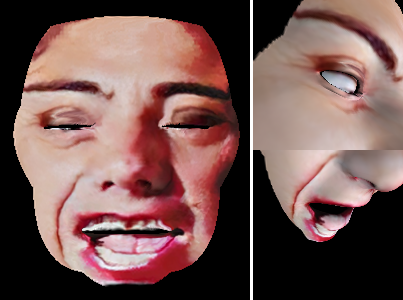}}
    \subcaptionbox*{(e)}{\includegraphics[width=0.11\linewidth]{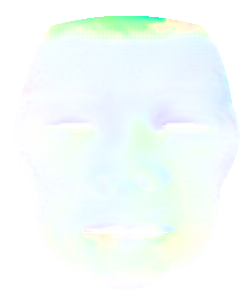}}
    \subcaptionbox*{(f)}{\includegraphics[width=0.17\linewidth]{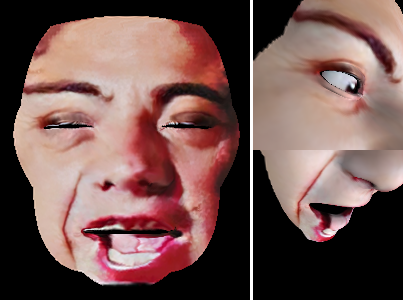}}
    
    \caption{Additional texture map refinement results on unseen real-world test images. (a)–(f) represent misaligned texture maps (baseline), texture maps with ARAP, rectified flow of TSAN, texture maps with TSAN, rectified flow of ARAP+TSAN, and texture maps with ARAP+TSAN, respectively. }
    \label{fig:more_qual_cmp_tsan}
\end{figure*}

\paragraph{More qualitative comparison results.} We provide more qualitative results to demonstrate the effectiveness of TSAN in~\cref{fig:more_qual_cmp_tsan}.

\begin{table*}[h]
    \centering
    \small
    \begin{tabular}{cccccccc}
        \toprule 
        \multirow{2}{*}{\makecell{Training \\ Dataset}} & \multirow{2}{*}{\makecell{Reconstruction \\ Method}} & \multirow{2}{*}{FER Model} & Config \#1 & Config \#2 & Config \#3 & Config \#4 & Config \#5 \\
        \cmidrule(lr){4-8} & & & UAR / WAR & UAR / WAR & UAR / WAR & UAR / WAR & UAR / WAR \\
 
        \midrule
        \multirow{2}{*}{RAF-DB} 
        & \multirow{2}{*}{None} & EfficientFace & 29.35 / 31.56 & 34.42 / 37.35 &  26.10 / 28.01 & 25.10 / 27.30  & 31.21 / 34.40 \\
        & & APViT & 26.32 / 30.50 & 26.19 / 29.43 & 21.05 / 24.59 &  27.37 / 31.56 & 26.19 / 29.43 \\
        
        \midrule
        \multirow{5}{*}{AffectNet-280K} & \multirow{5}{*}{None}
         
        & EfficientFace & 28.33 / 31.21 & 37.30 / 38.89 & 30.52 / 31.32 & 25.49 / 27.66 & 33.57 / 35.70 \\
        & & R-18 & 29.77 / 32.51 & 29.25 / 31.32 & 26.82 / 28.49 & 29.86 / 32.74 & 37.44 / 40.31 \\
        & & R-50 & 30.97 / 34.16 & 29.92 / 31.56 & 26.50 / 28.61 & 30.62 / 33.22 & 35.44 / 38.06 \\
        & & Swin-T & 42.22 / 44.80 & 37.37 / 38.77 & 32.50 / 34.04 & 30.37 / 32.86 & 40.90 / 43.97 \\
        & & Swin-B & \underline{42.57} / \underline{45.98} & 39.75 / 41.96 & 32.13 / 33.57 & 30.94 / 34.16 & 36.07 / 40.19 \\

        \midrule
        \multirow{2}{*}{AffectNet-7K} & \multirow{2}{*}{None} & R-18 & 26.40 / 24.66 & 27.91 / 27.30 & 17.62 / 14.54 & 15.63 / 12.17 & 21.36 / 18.20 \\ 
        & & Swin-T & 32.53 / 32.74 & 35.28 / 36.17 & 25.67 / 25.53 & 25.40 / 25.53 & 32.98 / 36.17 \\

        \midrule
        \multirow{4}{*}{SynHMC-7K} & DECA & \multirow{4}{*}{Swin-T} & 31.10 / 32.39  & 19.33 / 16.90 & 25.60 / 26.83 & 23.72 / 23.05 & 31.74 / 33.45 \\
        & EMOCA &  & 26.53 / 27.42 & 25.61 / 24.35 & 23.78 / 22.81 & 31.61 / 32.86 & 28.71 / 30.97 \\
        & FDS &  & 41.35 / 43.26 & 36.82 / 36.76 & \underline{44.38} / \underline{44.56} & 39.66 / 41.61 & 36.84 / 38.53 \\
        & FFHQ-UV &  & 34.12 / 35.11 & 31.77 / 32.03 & 40.33 / 42.79 & 32.35 / 33.22 & 40.06 / 42.79 \\

        \midrule
        \multirow{2}{*}{SynHMC-7K} & \multirow{2}{*}{SynHMC (Ours)} & R-18 & 42.10 / 44.33 & \underline{46.05} / \underline{48.94} & 41.37 / 43.97 & \underline{42.82} / \textbf{46.81} & \underline{42.55} / \underline{45.86} \\
        & & Swin-T & \textbf{45.44} / \textbf{48.23} & \textbf{48.05} / \textbf{51.30} & \textbf{45.06} / \textbf{48.23} & \textbf{43.08} / \underline{45.74} & \textbf{43.57} / \textbf{46.45} \\
    
        \bottomrule
    \end{tabular}
    \caption{
        \label{tab:ava256_scan_cfgs}
        Evaluation results on the \textbf{Ava256-Scan} dataset for each HMC configuration. "None" refers to training with frontal-view images.
    }
\end{table*}

\begin{table*}[h]
    \centering
    \small
    \begin{tabular}{cccccc}
        \toprule 
        \multirow{2}{*}{Training Dataset} & \multirow{2}{*}{Reconstruction Method} & \multirow{2}{*}{FER Model} & Config \#1 & Config \#3 & Config \#4 \\
        \cmidrule(lr){4-6} & & & Accuracy & Accuracy & Accuracy \\
 
        \midrule
        \multirow{2}{*}{RAF-DB} 
        & \multirow{2}{*}{None} & EfficientFace & 31.68  & 26.09  & 26.71 \\
        & & APViT & 31.68 & 19.25 & 21.74  \\
        
        \midrule
        \multirow{5}{*}{AffectNet-280K} & \multirow{5}{*}{None}
         
        & EfficientFace & 33.54 & 24.84 & 16.77 \\
        & & R-18 & 32.92 & 24.84 & 28.57 \\
        & & R-50 & 34.78 & 26.71 & 25.47 \\
        & & Swin-T & 39.75 & 31.68 & 37.27 \\
        & & Swin-B & 44.10 & 28.57 & 35.40 \\

        \midrule
        \multirow{2}{*}{AffectNet-7K} & \multirow{2}{*}{None} & R-18 & 26.09 & 19.25 & 21.74 \\
        & & Swin-T & 31.68 & 18.63 & 20.50 \\

        \midrule
        \multirow{4}{*}{SynHMC-7K} & DECA & \multirow{4}{*}{Swin-T} & 32.92  & 34.16  & 31.68 \\
        & EMOCA & & 45.34 & 34.16 & 38.51 \\
        & FDS & & 45.96 & \underline{45.96} & 44.10 \\
        & FFHQ-UV & & 45.34 & 44.10 & 38.51 \\

        \midrule
        \multirow{2}{*}{SynHMC-7K} & \multirow{2}{*}{SynHMC (Ours)} & R-18 & \underline{47.20} & 41.61 & \underline{44.72}  \\
        & & Swin-T & \textbf{47.83} & \textbf{50.31} & \textbf{46.58} \\
    
        \bottomrule
    \end{tabular}
    \caption{
        \label{tab:3drfe_cfgs}
        Evaluation results on the \textbf{3DRFE} dataset under HMC \emph{Config \#1} (four-view, FoV=66\textdegree), HMC \emph{Config \#3} (four-view, FoV=120\textdegree) and HMC \emph{Config \#4} (three-view, FoV=90\textdegree). "None" refers to training with frontal-view images.
    }
\end{table*}

\begin{figure*}[t]
    \centering
    \begin{tabular}{@{\hspace{0.1mm}}c@{\hspace{0.5mm}}c@{\hspace{1mm}}c@{\hspace{1mm}}c@{\hspace{1mm}}c@{}}
        \includegraphics[width=0.1535\textwidth]{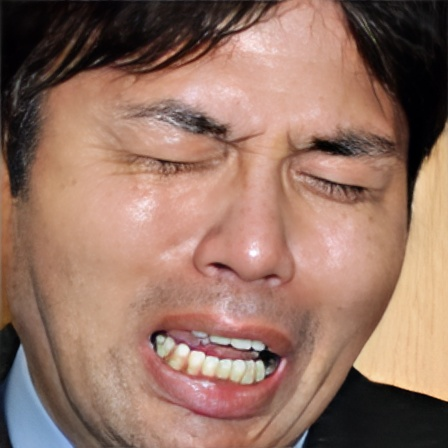} &
        \includegraphics[width=0.19\textwidth]{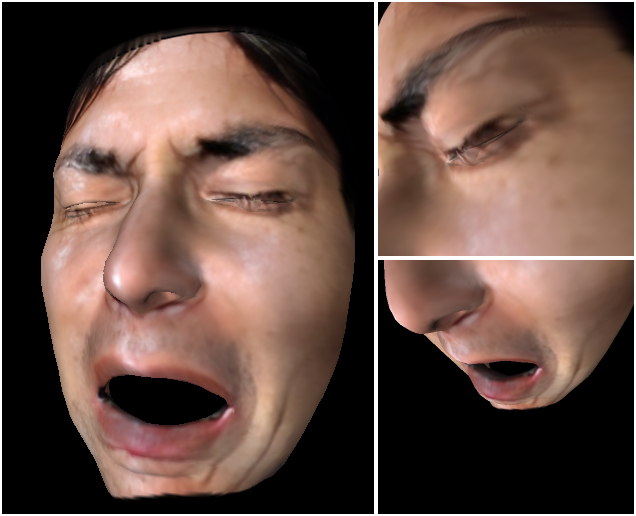} & \includegraphics[width=0.19\textwidth]{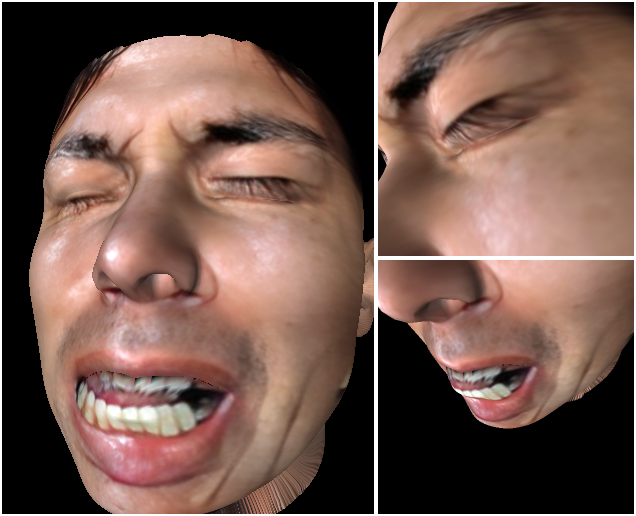}
        & \includegraphics[width=0.19\textwidth]{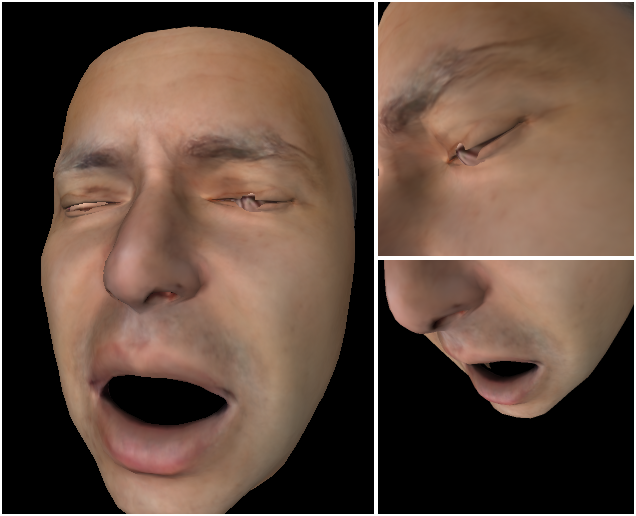} & \includegraphics[width=0.19\textwidth]{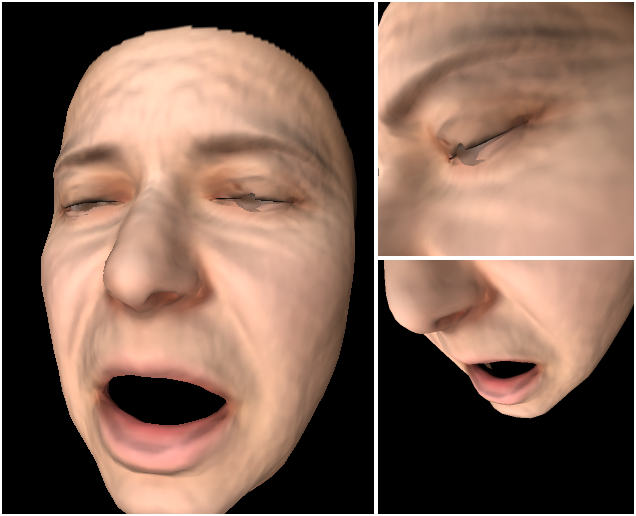} 
         \\

        \subcaptionbox*{Source Image}{\includegraphics[width=0.1535\textwidth]{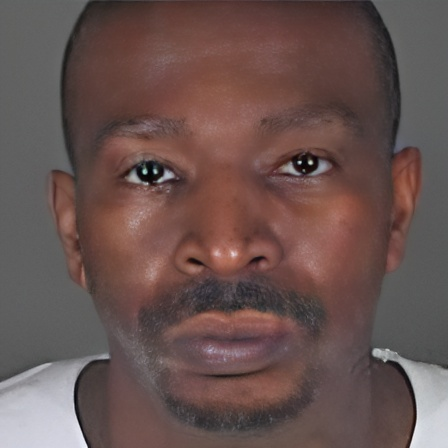}} & \subcaptionbox*{SynHMC (Ours)}{\includegraphics[width=0.19\textwidth]{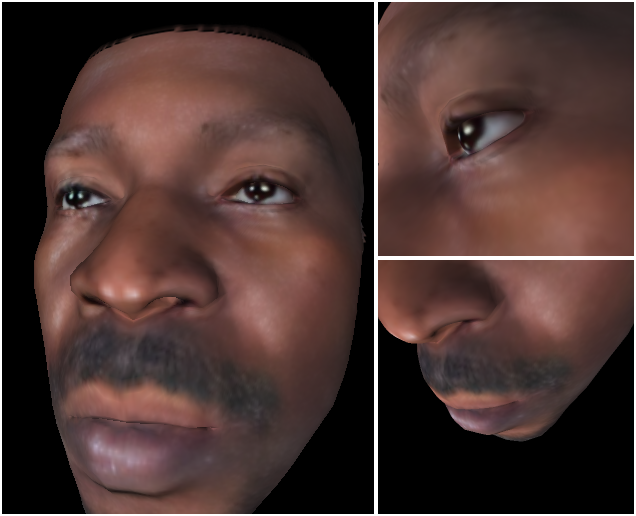}} & \subcaptionbox*{FDS}{\includegraphics[width=0.19\textwidth]{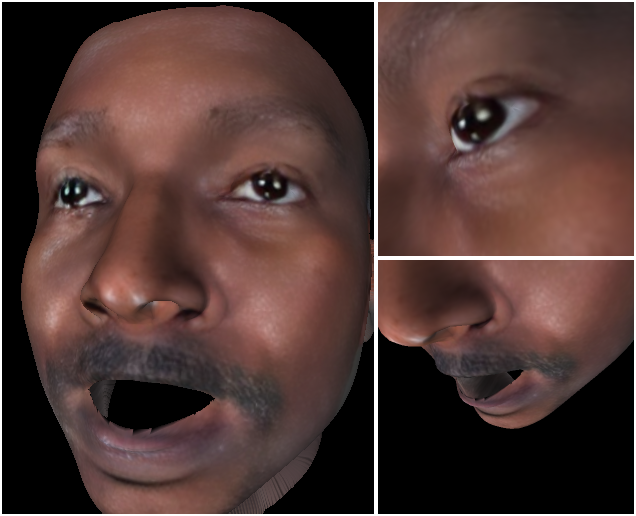}} & \subcaptionbox*{FFHQ-UV}{\includegraphics[width=0.19\textwidth]{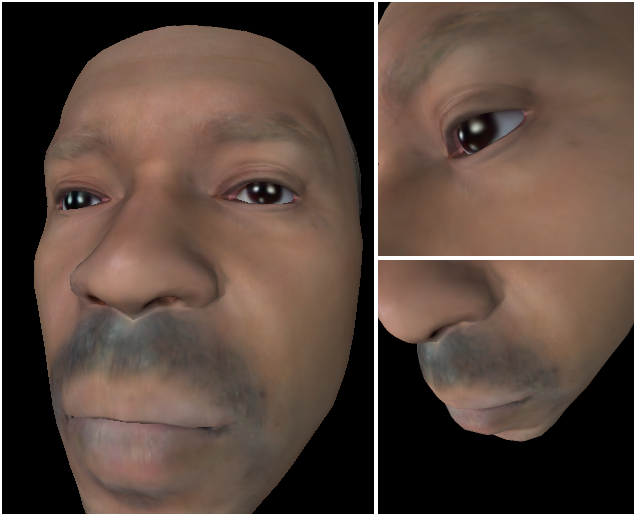}} & \subcaptionbox*{EMOCA}{\includegraphics[width=0.19\textwidth]{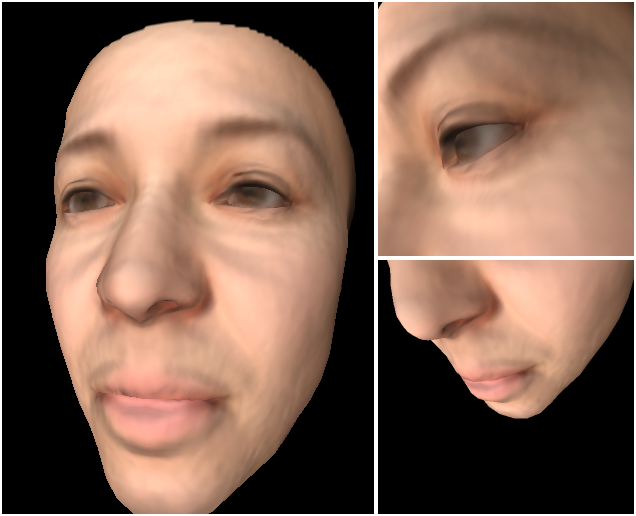}}
         \\
    \end{tabular}

    \caption{Comparison of 3D face reconstruction methods for HMC image synthesis. \emph{Config \#1} is used for rendering HMC images.}
    \label{fig:more_cmp_qual_recon_cfg1}
\end{figure*}

\begin{figure*}[t]
    \centering
    \begin{tabular}{@{\hspace{0.1mm}}c@{\hspace{0.5mm}}c@{\hspace{1mm}}c@{\hspace{1mm}}c@{\hspace{1mm}}c@{}}
        \includegraphics[width=0.1535\textwidth]{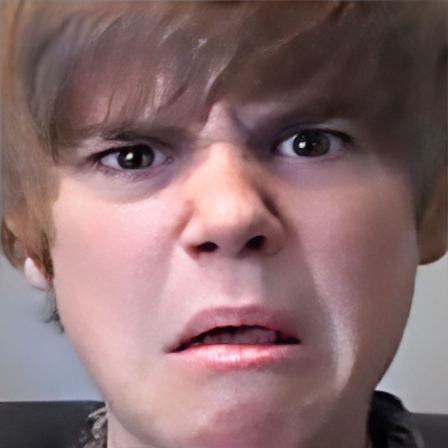} &
        \includegraphics[width=0.19\textwidth]{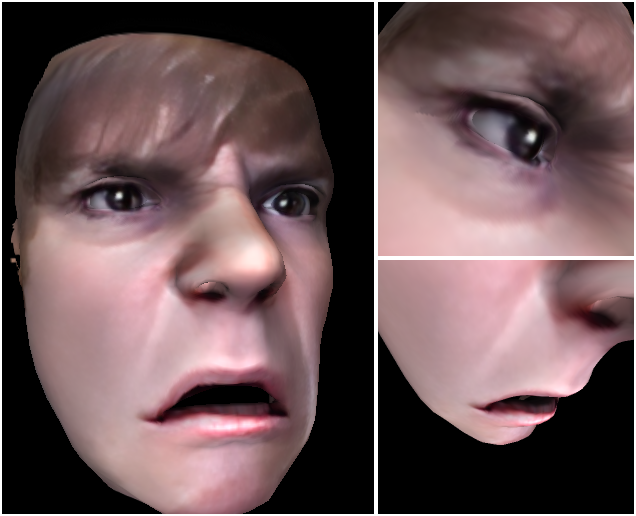} & \includegraphics[width=0.19\textwidth]{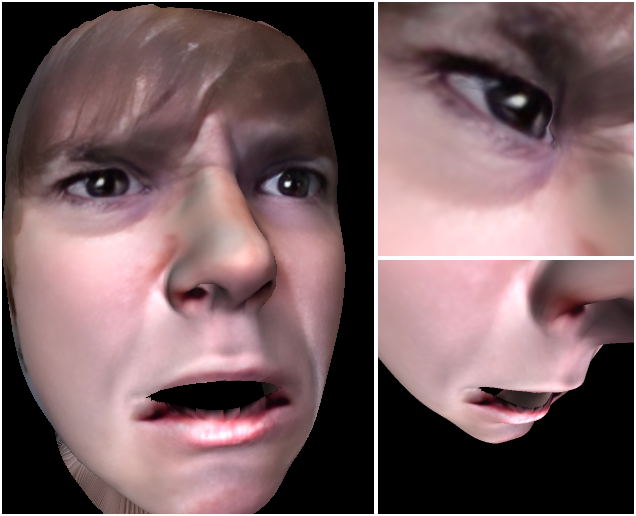}
        & \includegraphics[width=0.19\textwidth]{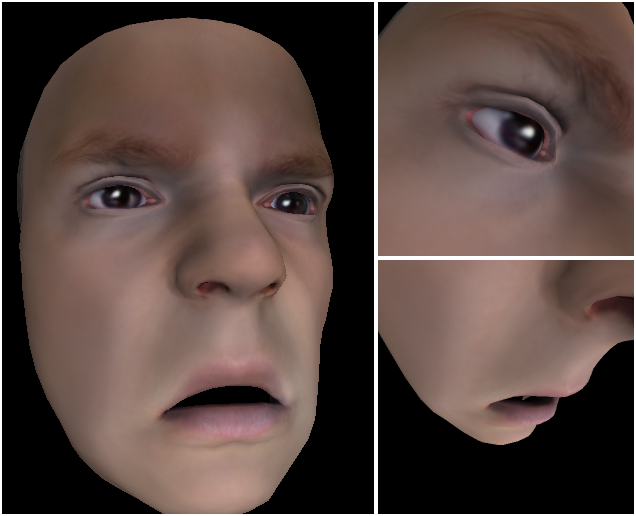} & \includegraphics[width=0.19\textwidth]{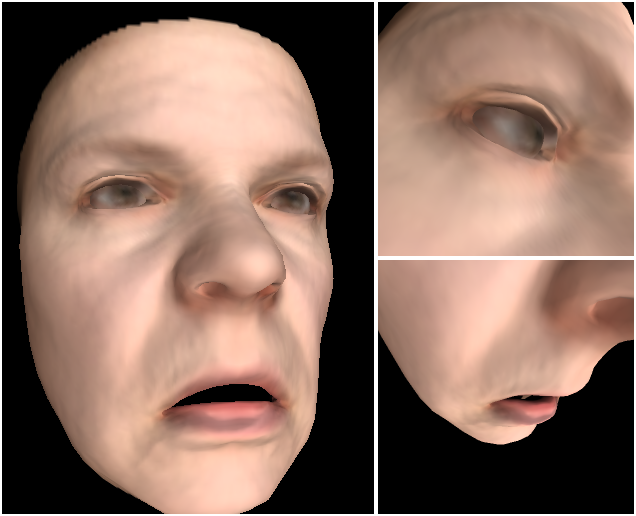} 
         \\

        \subcaptionbox*{Source Image}{\includegraphics[width=0.1535\textwidth]{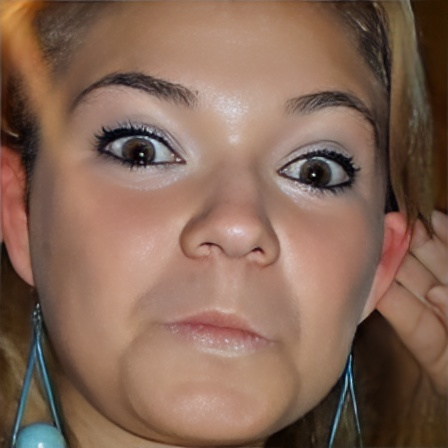}} & \subcaptionbox*{SynHMC (Ours)}{\includegraphics[width=0.19\textwidth]{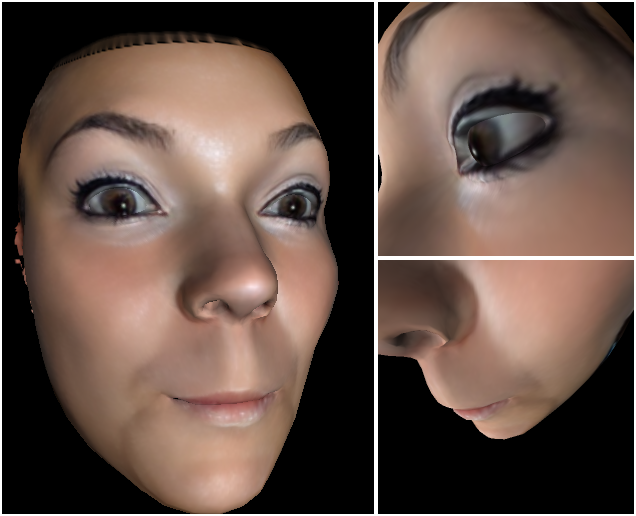}} & \subcaptionbox*{FDS}{\includegraphics[width=0.19\textwidth]{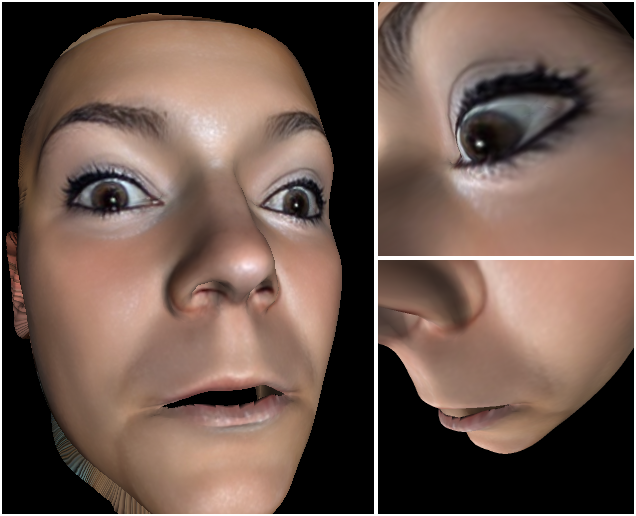}} & \subcaptionbox*{FFHQ-UV}{\includegraphics[width=0.19\textwidth]{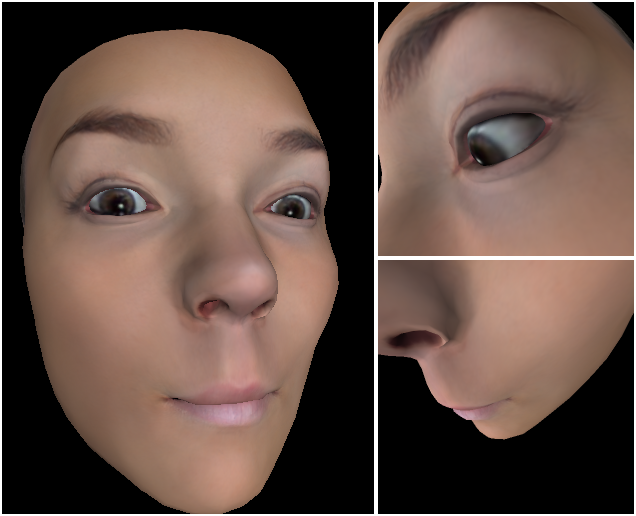}} & \subcaptionbox*{EMOCA}{\includegraphics[width=0.19\textwidth]{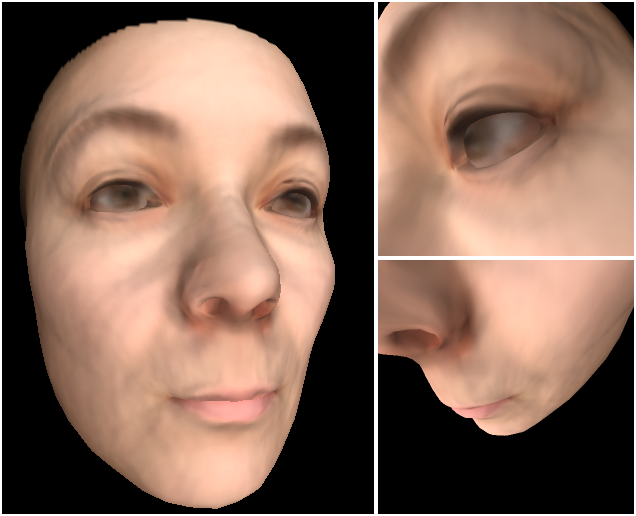}}
         \\
    \end{tabular}

    \caption{Comparison of 3D face reconstruction methods for HMC image synthesis. \emph{Config \#2} is used for rendering HMC images.}
    \label{fig:more_cmp_qual_recon_cfg2}
\end{figure*}

\begin{figure*}[t]
    \centering
    \begin{tabular}{@{\hspace{0.1mm}}c@{\hspace{0.5mm}}c@{\hspace{1mm}}c@{\hspace{1mm}}c@{\hspace{1mm}}c@{}}
        \includegraphics[width=0.1535\textwidth]{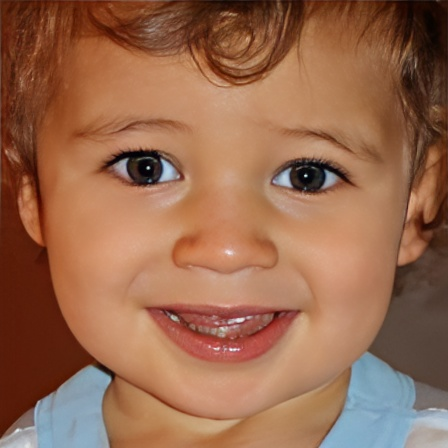} &
        \includegraphics[width=0.19\textwidth]{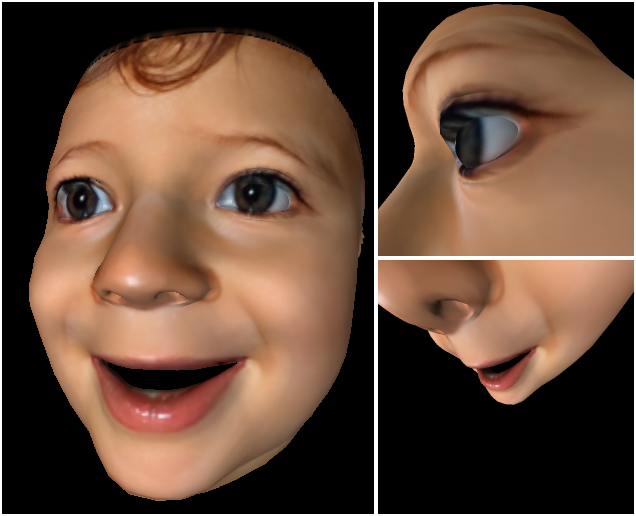} & \includegraphics[width=0.19\textwidth]{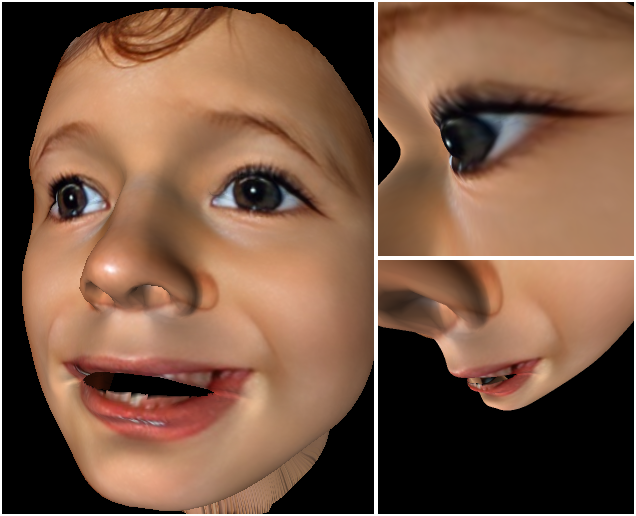}
        & \includegraphics[width=0.19\textwidth]{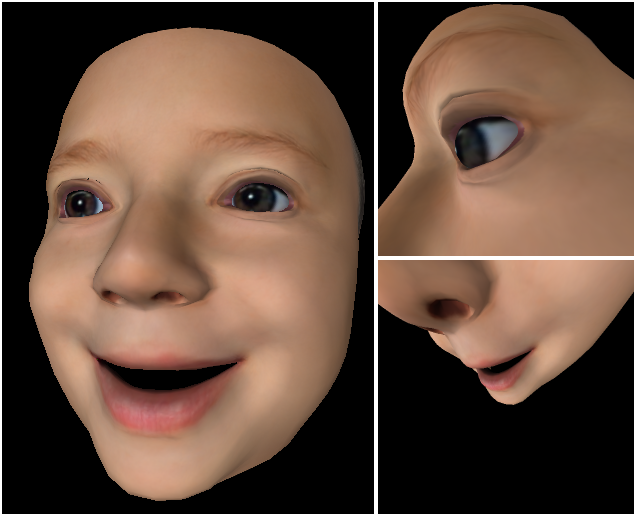} & \includegraphics[width=0.19\textwidth]{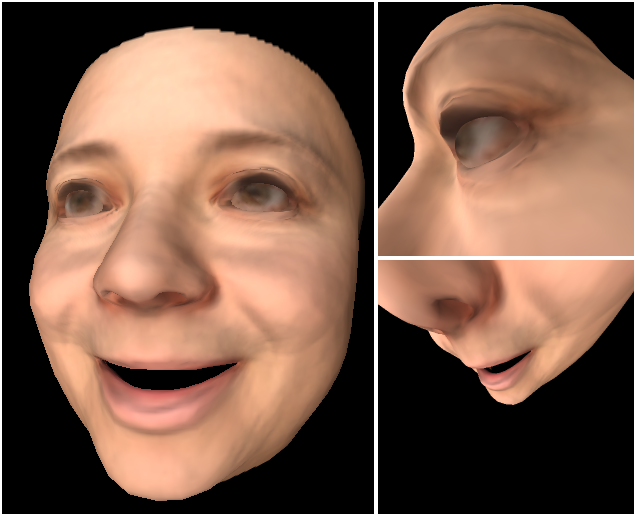} 
         \\

        \subcaptionbox*{Source Image}{\includegraphics[width=0.1535\textwidth]{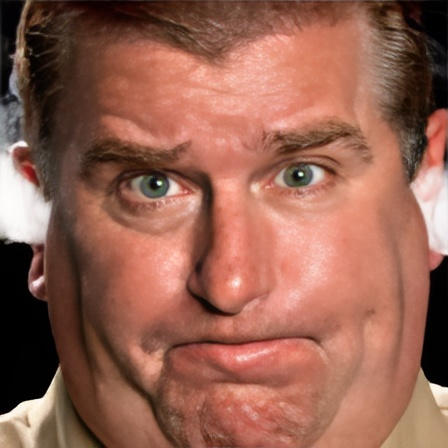}} & \subcaptionbox*{SynHMC (Ours)}{\includegraphics[width=0.19\textwidth]{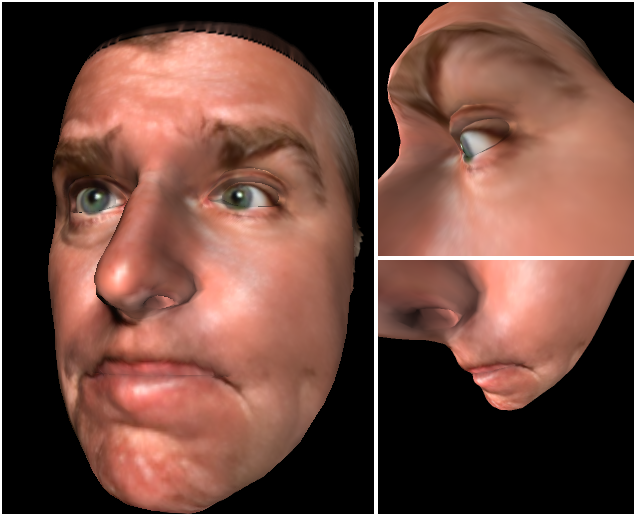}} & \subcaptionbox*{FDS}{\includegraphics[width=0.19\textwidth]{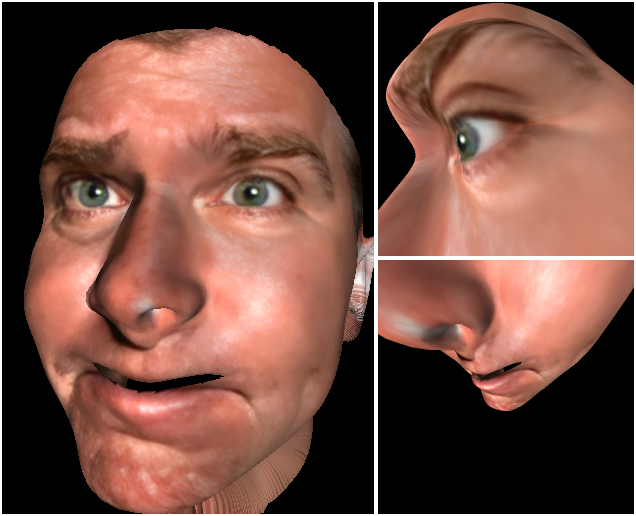}} & \subcaptionbox*{FFHQ-UV}{\includegraphics[width=0.19\textwidth]{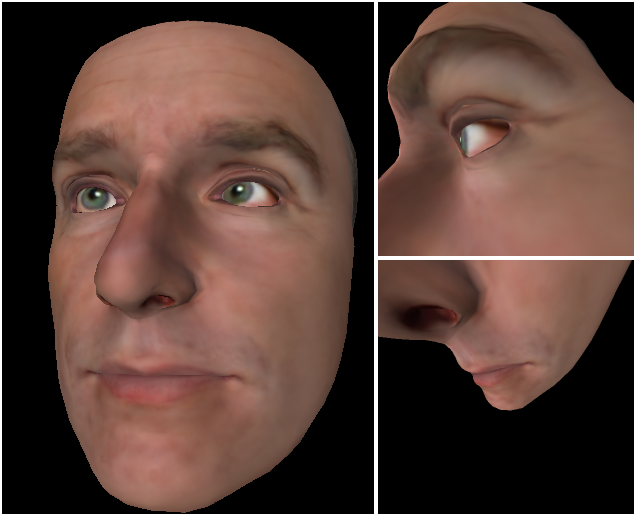}} & \subcaptionbox*{EMOCA}{\includegraphics[width=0.19\textwidth]{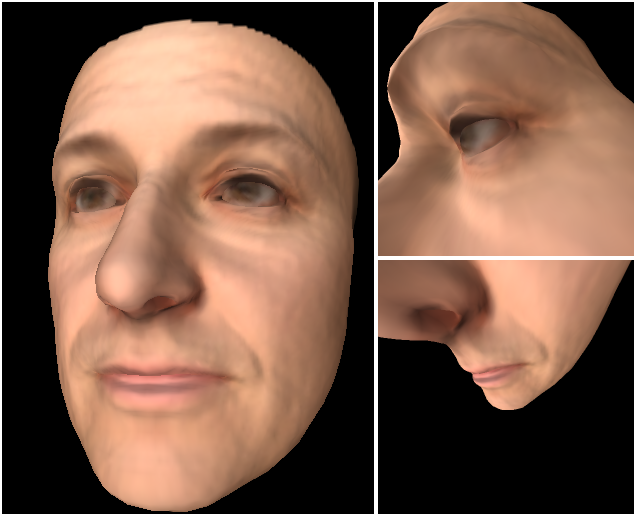}}
         \\
    \end{tabular}

    \caption{Comparison of 3D face reconstruction methods for HMC image synthesis. \emph{Config \#3} is used for rendering HMC images.}
    \label{fig:more_cmp_qual_recon_cfg3}
\end{figure*}

\begin{figure*}[t]
    \centering
    \begin{tabular}{@{\hspace{0.1mm}}c@{\hspace{0.5mm}}c@{\hspace{1mm}}c@{\hspace{1mm}}c@{\hspace{1mm}}c@{}}
        \includegraphics[width=0.1535\textwidth]{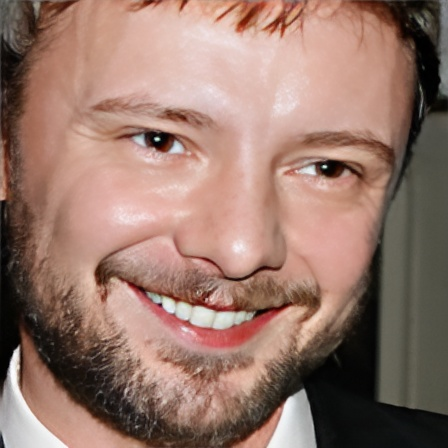} &
        \includegraphics[width=0.19\textwidth]{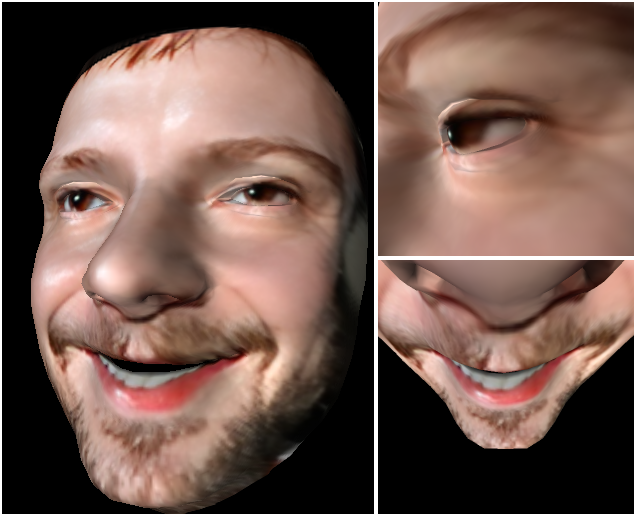} & \includegraphics[width=0.19\textwidth]{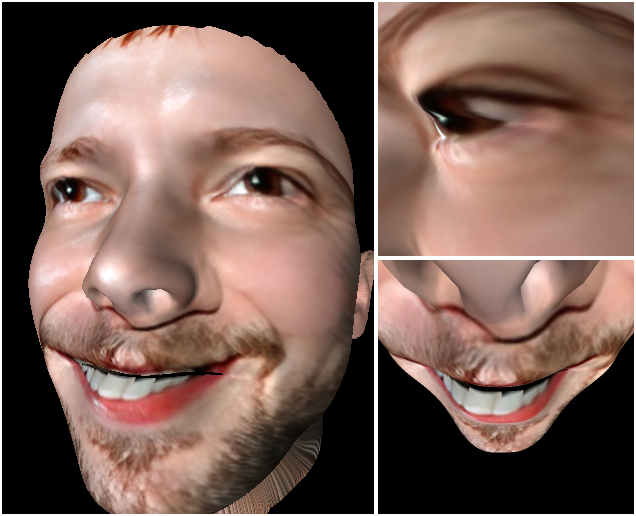}
        & \includegraphics[width=0.19\textwidth]{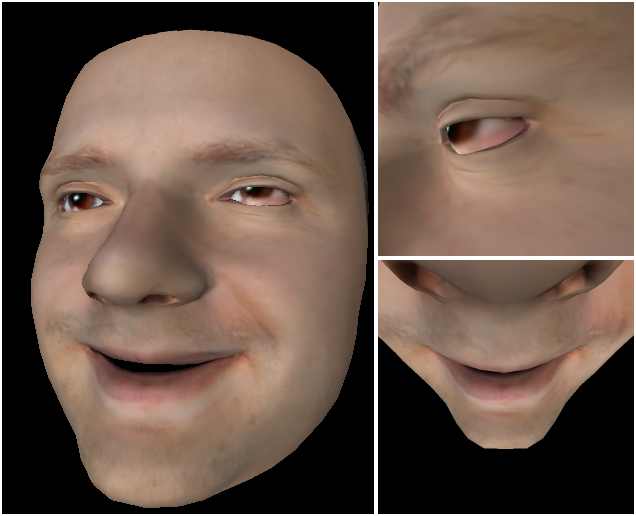} & \includegraphics[width=0.19\textwidth]{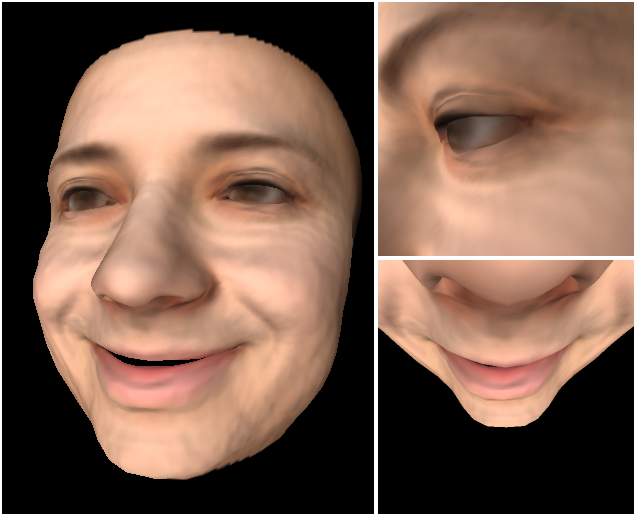} 
         \\

        \subcaptionbox*{Source Image}{\includegraphics[width=0.1535\textwidth]{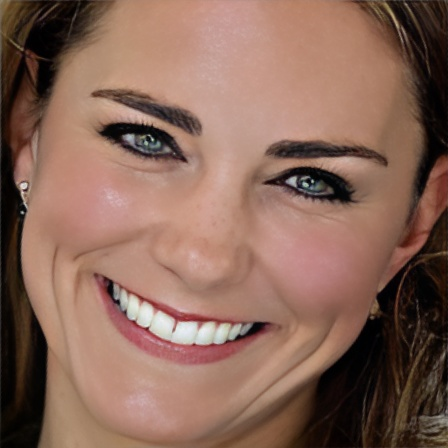}} & \subcaptionbox*{SynHMC (Ours)}{\includegraphics[width=0.19\textwidth]{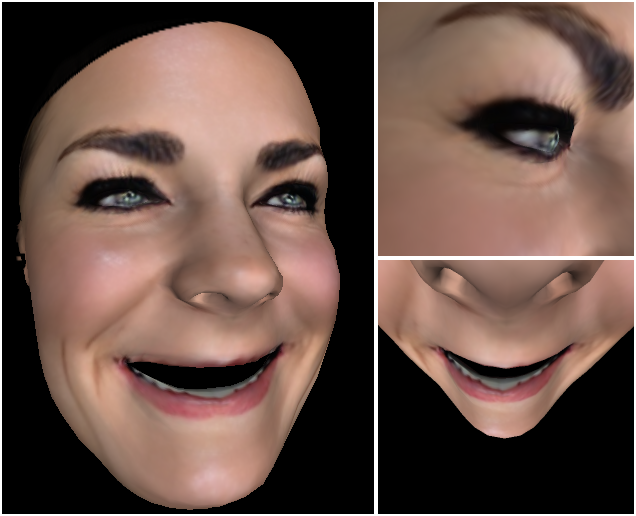}} & \subcaptionbox*{FDS}{\includegraphics[width=0.19\textwidth]{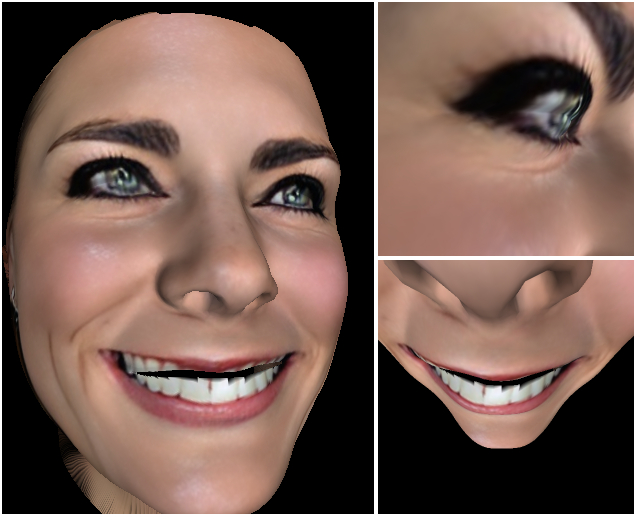}} & \subcaptionbox*{FFHQ-UV}{\includegraphics[width=0.19\textwidth]{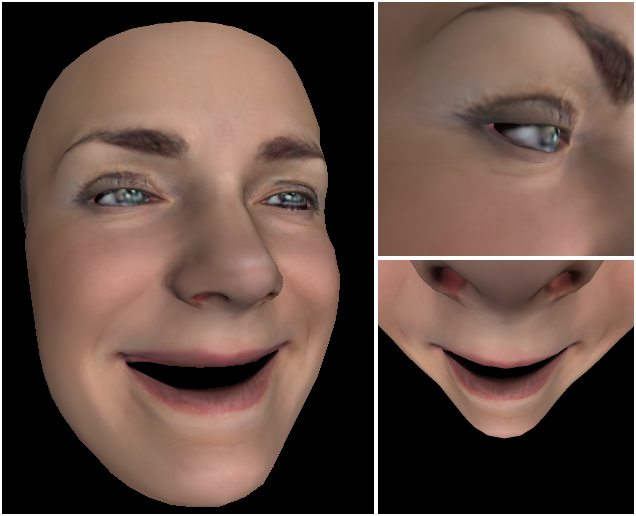}} & \subcaptionbox*{EMOCA}{\includegraphics[width=0.19\textwidth]{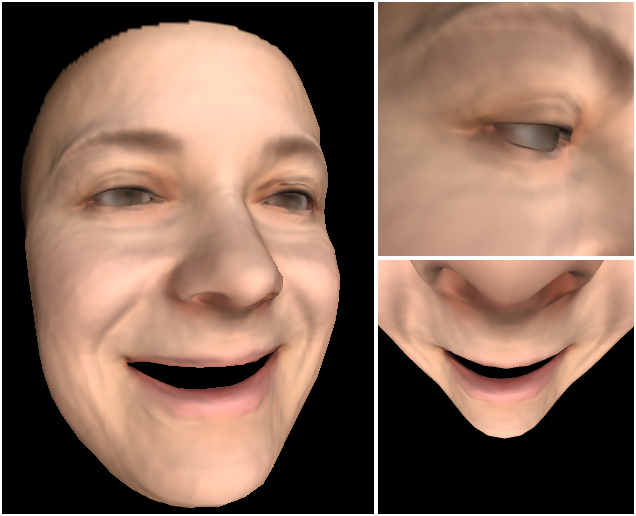}}
         \\
    \end{tabular}

    \caption{Comparison of 3D face reconstruction methods for HMC image synthesis. \emph{Config \#4} is used for rendering HMC images.}
    \label{fig:more_cmp_qual_recon_cfg4}
\end{figure*}

\begin{figure*}[t]
    \centering
    \begin{tabular}{@{\hspace{0.1mm}}c@{\hspace{0.5mm}}c@{\hspace{1mm}}c@{\hspace{1mm}}c@{\hspace{1mm}}c@{}}
        \includegraphics[width=0.1535\textwidth]{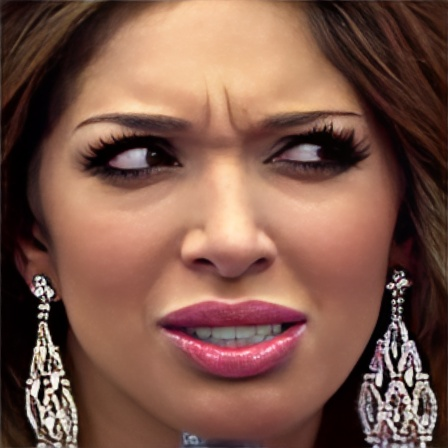} &
        \includegraphics[width=0.19\textwidth]{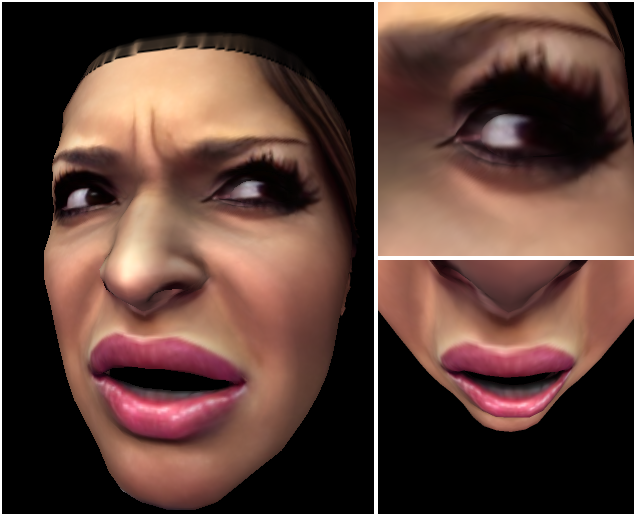} & \includegraphics[width=0.19\textwidth]{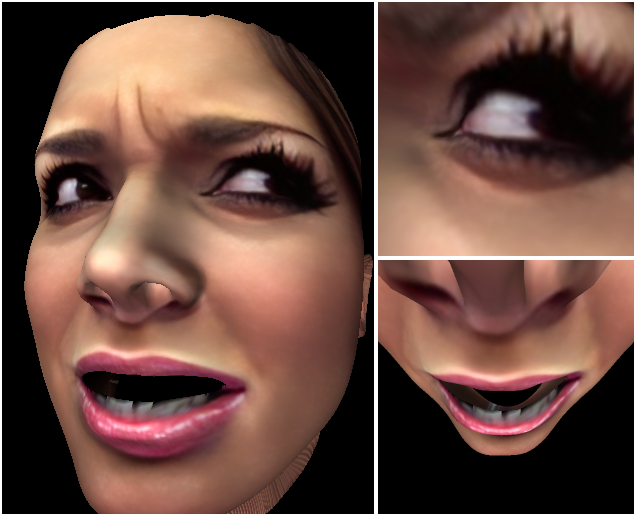}
        & \includegraphics[width=0.19\textwidth]{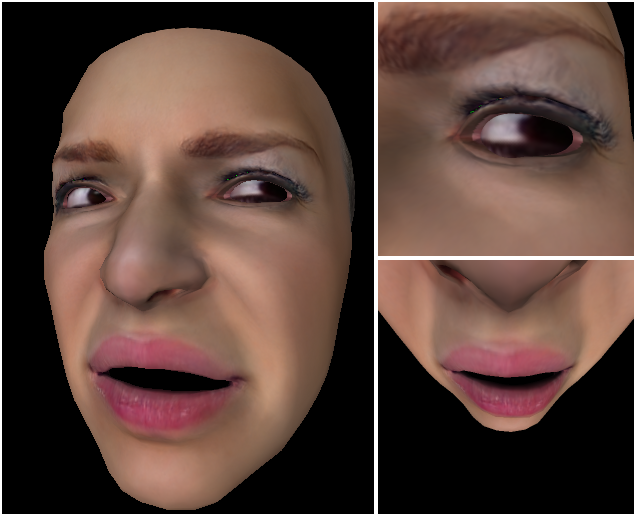} & \includegraphics[width=0.19\textwidth]{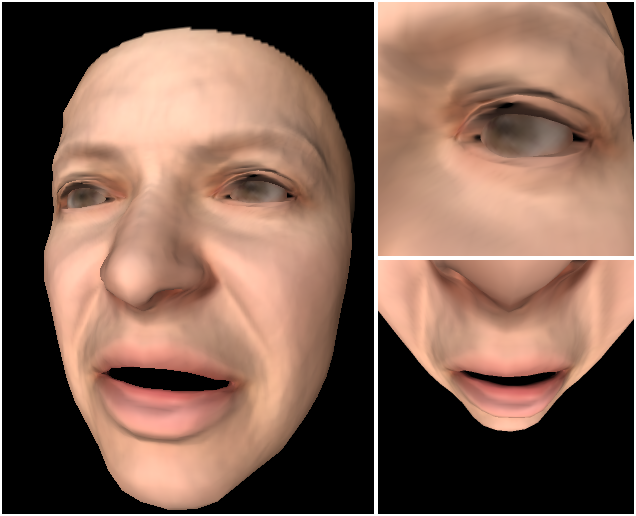} 
         \\

        \subcaptionbox*{Source Image}{\includegraphics[width=0.1535\textwidth]{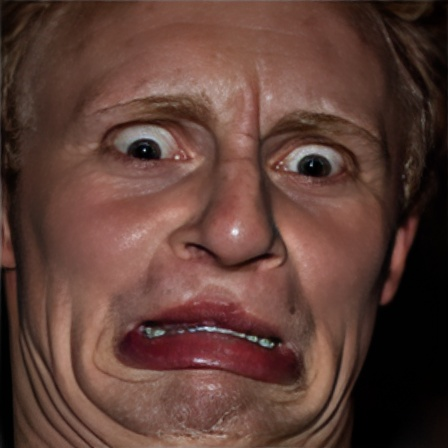}} & \subcaptionbox*{SynHMC (Ours)}{\includegraphics[width=0.19\textwidth]{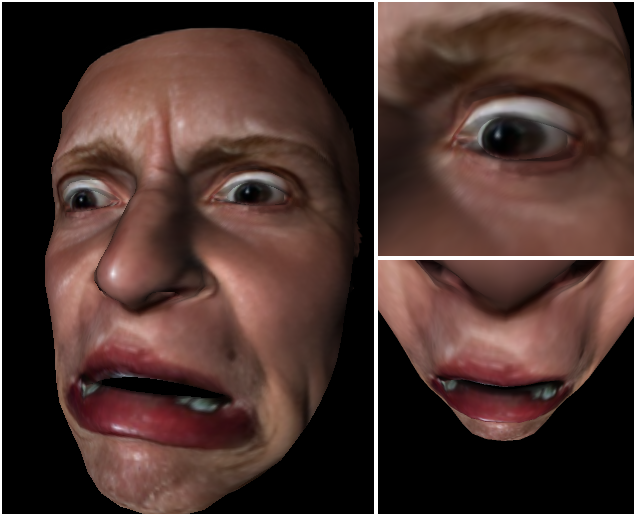}} & \subcaptionbox*{FDS}{\includegraphics[width=0.19\textwidth]{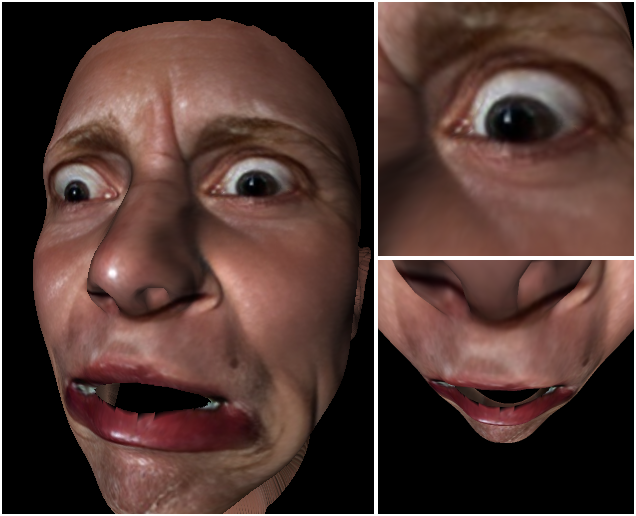}} & \subcaptionbox*{FFHQ-UV}{\includegraphics[width=0.19\textwidth]{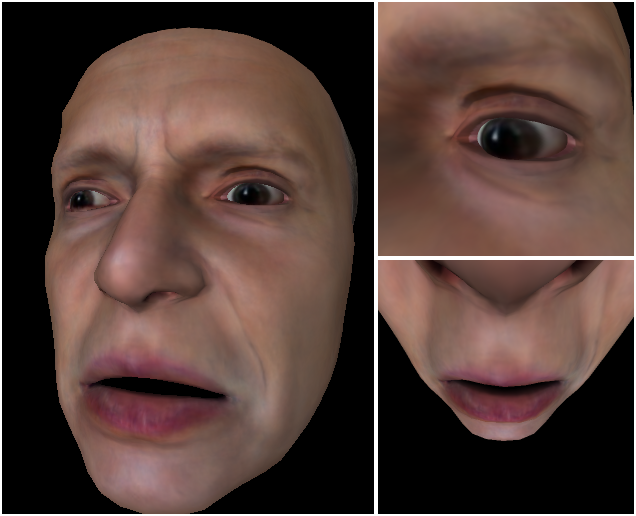}} & \subcaptionbox*{EMOCA}{\includegraphics[width=0.19\textwidth]{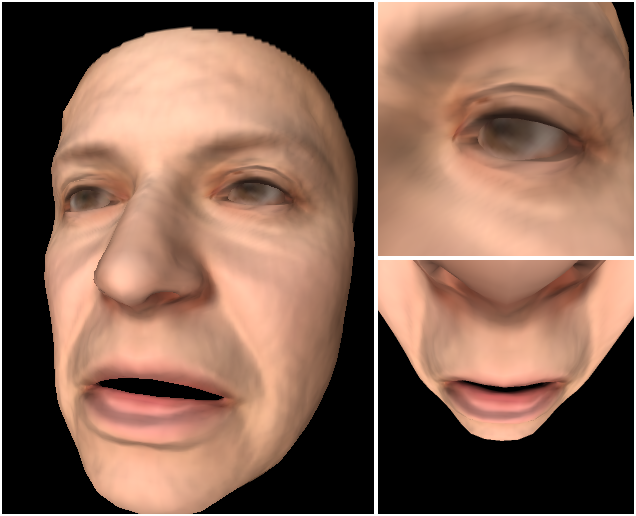}}
         \\
    \end{tabular}

    \caption{Comparison of 3D face reconstruction methods for HMC image synthesis. \emph{Config \#5} is used for rendering HMC images.}
    \label{fig:more_cmp_qual_recon_cfg5}
\end{figure*}

\section{Camera and Rendering Configurations}
\label{supp sec: camera system}

Following the camera setups in~\cite{martinezcodec} (4-view HMCs) and~\cite{jourabloo2022robust} (3-view HMCs), we employ five HMC configurations with similar visible facial areas in our experiments. The first three configurations share same camera views with similar visible facial regions, but have increasing FoVs of 66\textdegree, 90\textdegree, and 120\textdegree, respectively. The FoV parameters reference the Sony IMX708 sensor~\footnote{\url{https://www.raspberrypi.com/products/camera-module-3/}} and the OVM6211 sensor~\footnote{\url{https://www.ovt.com/products/ovm6211/}}.
We employ three-view HMCs in the last two configurations under FoV of 90\textdegree, with the main differences in the eye camera placements. Config \#4 features side eye cameras, while Config \#5 positions them more frontally.
We implement the HMCs using PyTorch3D's \texttt{FoVPerspectiveCameras}. Detailed camera parameters of each configuration are provided in~\cref{tab:cam_cfg}.

We use PyTorch3D's \texttt{MeshRasterizer} to render HMC images, with lighting intensities of 1.5 for AffectNet and 3DRFE dataset and 2.5 for Ava256. We set the rendering resolution to $512\times 512$ for all images.

\section{Ava256 Dataset}
\label{supp sec: ava256}

\begin{figure}[h]
    \centering
    \includegraphics[draft=false, width=0.95\linewidth]{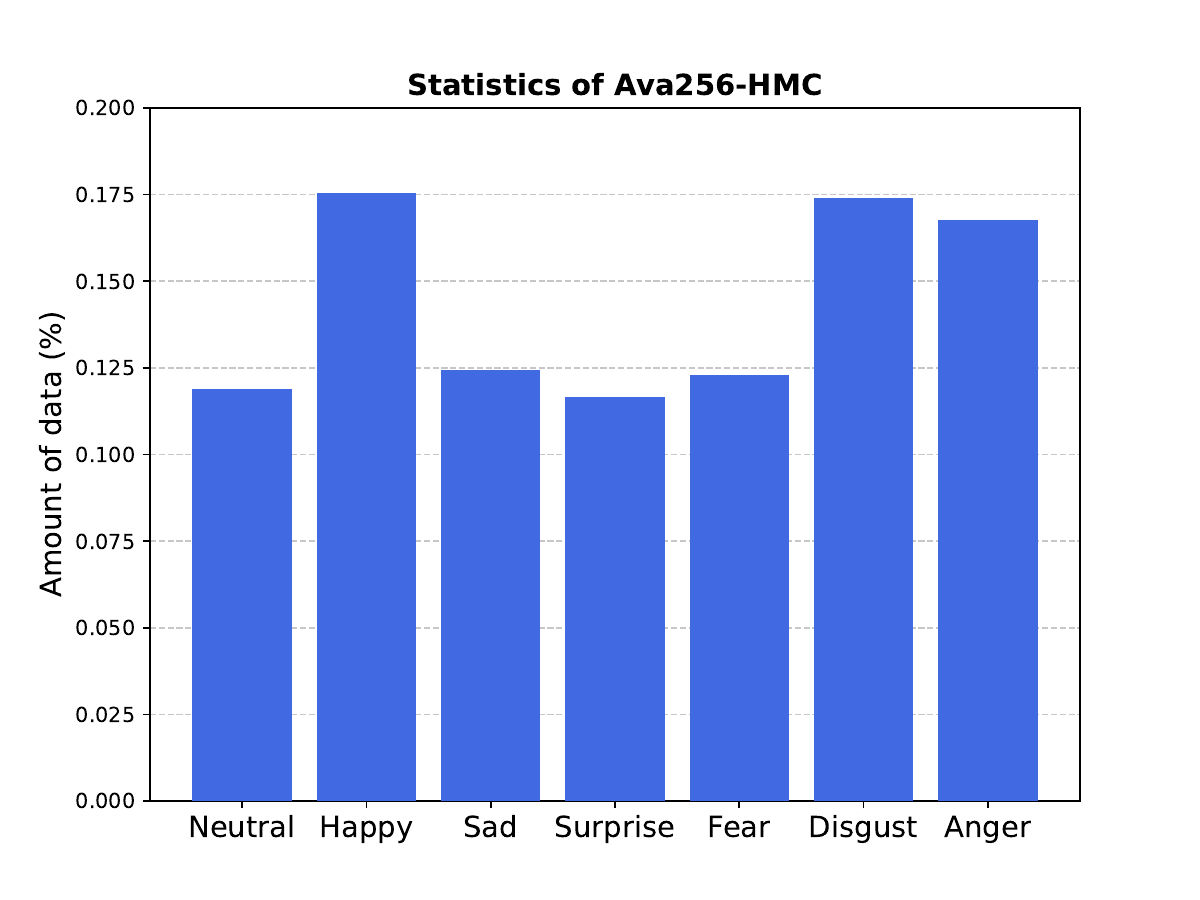}

    \caption{Statistics of the Ava256-HMC dataset for seven-class FER after grouping subclasses.}
    \label{fig:stat_ava256_hmc}
\end{figure}

\paragraph{Grouping sub-classes for FER.} Since the original Ava256 clip annotations do not follow the seven basic expressions, we regroup them according to~\cref{tab:ava256_regroup} for seven-class FER. \Cref{fig:stat_ava256_hmc} provides the statistics of the resulting  Ava256-HMC datasets.

\begin{figure*}[t]
    \centering
    \includegraphics[draft=false, width=0.95\linewidth]{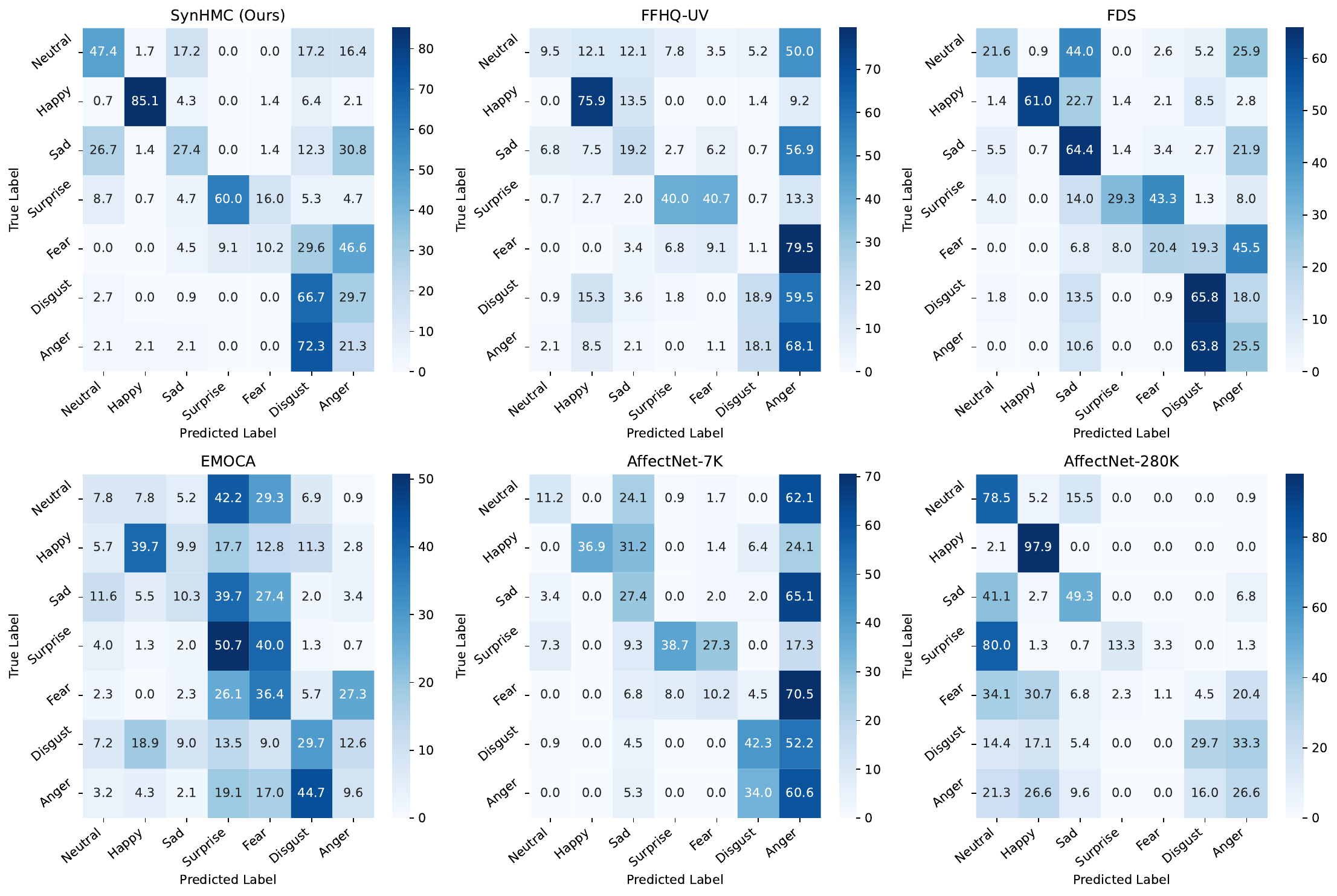}

    \caption{Confusion matrices of different methods on \textbf{Ava256-Scan} on \emph{Config \#1}, with Swin-T  as the FER model for all methods.}
    \label{fig:cm}
\end{figure*}

\paragraph{More comparisons results.} We provide the performance of models on each HMC configuration in~\cref{tab:ava256_scan_cfgs} and provide the confusion matrices under config \#1 in ~\cref{fig:cm}. ~\cref{fig:more_cmp_qual_recon_cfg1}, ~\cref{fig:more_cmp_qual_recon_cfg2}, ~\cref{fig:more_cmp_qual_recon_cfg3}, ~\cref{fig:more_cmp_qual_recon_cfg4}, ~\cref{fig:more_cmp_qual_recon_cfg5} offer additional visual comparisons of different 3D face reconstruction methods under each HMC configuration.

\begin{table}[h]
    \centering
    \small
    \begin{tabular}{ll}
        \toprule 
        Basic Expression & Ava256 Clip Label \\
        
        \midrule
        \multirow{5}{*}{Neutral} & neutral  \\
        & EXP\_eye\_neutral \\
        & EXP\_neutral\_peak \\
        & neutral\_comfortable \\
        & neutral\_guided \\

        \midrule
        \multirow{5}{*}{Happy} & large\_smile\_1  \\
        & mouth\_smile\_cheek\_raise \\
        & emotion\_images\_amusement \\
        & emotion\_images\_ecstasy \\
        & N\_EMO\_excitement\_joy \\

        \midrule
        \multirow{5}{*}{Sad} & sadness  \\
        & EMO\_sadness \\
        & EMO\_sad \\
        & emotion\_image\_sadness \\
        & emotion\_sentences\_sadness \\

        \midrule
        \multirow{3}{*}{Surprise} & surprise  \\
        & raised\_eyebrows \\
        & EMO\_amazement \\

        \midrule
        \multirow{6}{*}{Fear} & fear  \\
        & fear\_eyebrows \\
        & fear\_mouth \\
        & EMO\_fear \\
        & EMO\_terror \\
        & EMO\_apprehension \\

        \midrule
        \multirow{5}{*}{Disgust} & disgust\_1  \\
        & disgust\_2 \\
        & EMO\_disgust \\
        & emotion\_image\_disgust \\
        & emotion\_sentences\_disgust \\

        \midrule
        \multirow{6}{*}{Anger} & anger  \\
        & intense\_stare \\
        & brows\_lowered\_eyes\_wide \\
        & EMO\_anger \\
        & emotion\_image\_anger \\
        & emotion\_sentences\_anger \\
        
        \bottomrule
    \end{tabular}
    \caption{
        \label{tab:ava256_regroup}
        Subclass grouping for seven-class FER in the Ava256 dataset.
    }
\end{table}

\begin{table*}[h]
    \centering
    \scriptsize
    
    \begin{tabular}{l|c|c|c}
        \toprule
        & Config \#1 & Config \#2  & Config \#3 \\

        \midrule
        \textbf{FoV} & 66 & 90 & 120 \\
        
        $\mathbf{T}_\mathrm{right\_eye}$ & 
        $\begin{bmatrix} 
            -0.601 & -0.394 &  0.695 & 0.004 \\ 
            -0.421 &  0.896 &  0.144 & -0.029 \\ 
            -0.679 & -0.207 & -0.704 & 0.094 \\
        \end{bmatrix}$ &
        $\begin{bmatrix} 
            -0.663 & -0.428 &  0.615 & 0.003 \\ 
            -0.428 &  0.890 &  0.159 & -0.031 \\ 
            -0.615 & -0.1578 & -0.773 & 0.073 \\
        \end{bmatrix}$ &
        $\begin{bmatrix} 
            -0.636 & -0.404 &  0.657 & 0.001 \\ 
            -0.421 &  0.896 &  0.144 & -0.031 \\ 
            -0.647 & -0.186 & -0.740 & 0.065 \\
        \end{bmatrix}$ \\

        $\mathbf{T}_\mathrm{left\_eye}$& 
        $\begin{bmatrix} 
            -0.601 &  0.394 & -0.695 & -0.004 \\ 
            0.421 &  0.896 &  0.143 & -0.029 \\ 
            0.679 &-0.207 & -0.704 & 0.094 \\
        \end{bmatrix}$ 
        & 
        $\begin{bmatrix} 
            -0.663 &  0.428 & -0.615 & -0.003 \\ 
            0.428 &  0.890 &  0.159 & -0.031 \\ 
            0.615 & -0.158 & -0.773 & 0.073 \\
        \end{bmatrix}$ &
        $\begin{bmatrix} 
            -0.636 &  0.404 & -0.657 & -0.001 \\ 
            0.421 &  0.896 &  0.144 & -0.031 \\ 
            0.647 & -0.186 & -0.740 & 0.065 \\
        \end{bmatrix}$ 
        \\

        $\mathbf{T}_\mathrm{right\_mouth}$&
        $\begin{bmatrix} 
            -0.862 &  0.088 &  0.500 & -0.005 \\ 
            -0.298 &  0.711 & -0.637 & 0.076 \\ 
            -0.412 & -0.698 & -0.587 & 0.101 \\
        \end{bmatrix}$ 
        &
        $\begin{bmatrix} 
            -0.860 & 0.055 & 0.508 & -0.008 \\ 
            -0.313 &  0.728 & -0.609 & 0.066 \\ 
            -0.404 & -0.683 & -0.609 & 0.063 \\
        \end{bmatrix}$ 
        &
        $\begin{bmatrix} 
            -0.862 &  0.088 & 0.500 & -0.005 \\
            -0.298 &  0.711 & -0.637 & 0.076 \\
            -0.412 & -0.698 & -0.587 & 0.056
        \end{bmatrix}$ 
        \\

        $\mathbf{T}_\mathrm{left\_mouth}$ &
        $\begin{bmatrix} 
            -0.862 &  -0.088 &  -0.500 & 0.005 \\ 
            0.298 &  0.711 & -0.637 & 0.076 \\ 
            0.412 & -0.698 & -0.587 & 0.101 \\
        \end{bmatrix}$  
        &
        $\begin{bmatrix} 
            -0.860 & -0.055 & -0.508 & 0.008 \\ 
            0.313 &  0.728 & -0.609 & 0.066 \\ 
            0.404 & -0.683 & -0.609 & 0.063 \\
        \end{bmatrix}$ 
        &
        $\begin{bmatrix} 
            -0.862 &  -0.088 & -0.500 & 0.01 \\
            0.298 &  0.711 & -0.637 & 0.076 \\
            0.412 & -0.698 & -0.587 & 0.056
        \end{bmatrix}$ 
        \\

        \toprule
        & Config \#4 & Config \#5 & \\
        \midrule

        \textbf{FoV} & 90 & 90 \\
        $\mathbf{T}_\mathrm{right\_eye}$ & 
        $\begin{bmatrix} 
            -0.688 & -0.162 & 0.708 & -0.003 \\
            -0.266 &  0.963 & -0.037 & -0.021 \\
            -0.676 & -0.214 & -0.706 & 0.078
        \end{bmatrix}$ 
        & 
        $\begin{bmatrix} 
            -0.916 &  0.0000 & 0.402 & -0.015 \\
            0.0000 &  1.0000 &  0.0000 & -0.025 \\
            -0.402 &  0.0000 & -0.916 & 0.071
        \end{bmatrix}$ 
        \\

        $\mathbf{T}_\mathrm{left\_eye}$ & 
        $\begin{bmatrix} 
            -0.688 & 0.162 & -0.708 & 0.003 \\
            0.266 &  0.963 & -0.037 & -0.021 \\
            0.676 & -0.214 & -0.706 & 0.078
        \end{bmatrix}$ 
        & 
        $\begin{bmatrix} 
            -0.916 &  0.0000 & -0.402 & 0.015 \\
            0.0000 &  1.0000 &  0.0000 & -0.025 \\
            0.402 &  0.0000 & -0.916 & 0.071
        \end{bmatrix}$ 
        \\

        $\mathbf{T}_\mathrm{top\_mouth}$ & 
        $\begin{bmatrix} 
            -1.000 & 0.000 & 0.000 & 0.000 \\
            0.000 & 0.766 & -0.643 & 0.067 \\
            0.000 & -0.643 & -0.766 & 0.061
        \end{bmatrix}$ 
        & 
        $\begin{bmatrix} 
            -1.000 & 0.000 & 0.000 & 0.000 \\
            0.000 & 0.766 & -0.643 & 0.067 \\
            0.000 & -0.643 & -0.766 & 0.061
        \end{bmatrix}$  
        \\
        \bottomrule
    \end{tabular}
    \caption{HMC configurations parameterized in PyTorch3D's camera system, where $\mathbf{T}$ is the world-to-camera transformation matrix.}
    \label{tab:cam_cfg}
\end{table*}

\begin{table*}[t]
    \centering
    \begin{tabular}{|c|c|c|c|}
        \hline
        \textbf{Stage} & \textbf{Layer Type} & \textbf{Channels} & \textbf{Output Size (H×W)} \\
        \hline
        \textbf{Input} & - & 3 (RGB) & 256 × 256 \\
        \hline
        Conv & Conv (3×3) & 64 & 256 × 256 \\
        \textbf{Pos Encoding} & Concat & 64 & 256 × 256 \\
        \hline
        \multicolumn{4}{|c|}{\textbf{Contracting Path (Encoder)}} \\
        \hline
        Block 1 & Conv (3×3) + ReLU & 64 & 256 × 256 \\
        & Conv (3×3) + ReLU & 64 & 256 × 256 \\
        & Max Pooling (2×2) & - & 128 × 128 \\
        \hline
        Block 2 & Conv (3×3) + ReLU & 128 & 128 × 128 \\
        & Conv (3×3) + ReLU & 128 & 128 × 128 \\
        & Max Pooling (2×2) & - & 64 × 64 \\
        \hline
        Block 3 & Conv (3×3) + ReLU & 256 & 64 × 64 \\
        & Conv (3×3) + ReLU & 256 & 64 × 64 \\
        & Max Pooling (2×2) & - & 32 × 32 \\
        \hline
        Block 4 & Conv (3×3) + ReLU & 512 & 32 × 32 \\
        & Conv (3×3) + ReLU & 512 & 32 × 32 \\
        & Max Pooling (2×2) & - & 16 × 16 \\
        \hline
        \textbf{Bottleneck} & Conv (3×3) + ReLU & 1024 & 16 × 16 \\
        & Conv (3×3) + ReLU & 1024 & 16 × 16 \\
        \hline
        \multicolumn{4}{|c|}{\textbf{Expanding Path (Decoder)}} \\
        \hline
        Block 1 & Transposed Conv (2×2) & 512 & 32 × 32 \\
        & Skip Connection & 1024 & 32 × 32 \\
        & Conv (3×3) + ReLU & 512 & 32 × 32 \\
        & Conv (3×3) + ReLU & 512 & 32 × 32 \\
        \hline
        Block 2 & Transposed Conv (2×2) & 256 & 64 × 64 \\
        & Skip Connection & 512 & 64 × 64 \\
        & Conv (3×3) + ReLU & 256 & 64 × 64 \\
        & Conv (3×3) + ReLU & 256 & 64 × 64 \\
        \hline
        Block 3 & Transposed Conv (2×2) & 128 & 128 × 128 \\
        & Skip Connection & 256 & 128 × 128 \\
        & Conv (3×3) + ReLU & 128 & 128 × 128 \\
        & Conv (3×3) + ReLU & 128 & 128 × 128 \\
        \hline
        Block 4 & Transposed Conv (2×2) & 64 & 256 × 256 \\
        & Skip Connection & 128 & 256 × 256 \\
        & Conv (3×3) + ReLU & 64 & 256 × 256 \\
        & Conv (3×3) + ReLU & 64 & 256 × 256 \\
        \hline
        \textbf{Output} & Conv (1×1) & 2 & 256 × 256 \\
        \hline
    \end{tabular}
    \caption{Architecture of the TSAN for rectified flow estimation.}
    \label{tab:tsan_arch}
\end{table*}

\end{document}